\documentclass[10pt,twocolumn,letterpaper]{article}
\usepackage{multirow}

\usepackage[table]{xcolor} 
\usepackage{graphicx}
\usepackage{xcolor}       
\usepackage{booktabs}
\usepackage{amsmath}
\usepackage{graphicx}   
\usepackage{array}      
\usepackage{tabularx}
\usepackage{tcolorbox}
\usepackage{algorithm}
\usepackage{algpseudocode}
\usepackage{color, colortbl}
\usepackage{listings}

\usepackage{etoolbox}
\makeatletter
\AfterEndEnvironment{algorithm}{\let\@algcomment\relax}
\AtEndEnvironment{algorithm}{\kern2pt\hrule\relax\vskip3pt\@algcomment}
\let\@algcomment\relax
\newcommand\algcomment[1]{\def\@algcomment{\footnotesize#1}}
\renewcommand\fs@ruled{\def\@fs@cfont{\bfseries}\let\@fs@capt\floatc@ruled
  \def\@fs@pre{\hrule height.8pt depth0pt \kern2pt}%
  \def\@fs@post{}%
  \def\@fs@mid{\kern2pt\hrule\kern2pt}%
  \let\@fs@iftopcapt\iftrue}
\makeatother

\usepackage{cvpr}              
\definecolor{cvprblue}{rgb}{0.21,0.49,0.74}
\usepackage[pagebackref,breaklinks,colorlinks,allcolors=cvprblue]{hyperref}

\def\paperID{}
\def\confName{CVPR}
\def\confYear{2026}

\newlength\savewidth\newcommand\shline{\noalign{\global\savewidth\arrayrulewidth
  \global\arrayrulewidth 1pt}\hline\noalign{\global\arrayrulewidth\savewidth}}

\newcommand\midline{\noalign{\global\savewidth\arrayrulewidth
  \global\arrayrulewidth 0.5pt}\hline\noalign{\global\arrayrulewidth\savewidth}}

\newcommand{\tablestyle}[2]{\setlength{\tabcolsep}{#1}\renewcommand{\arraystretch}{#2}\centering\footnotesize}

\newcolumntype{x}[1]{>{\centering\arraybackslash}p{#1pt}}
\newcolumntype{y}[1]{>{\raggedright\arraybackslash}p{#1pt}}
\newcolumntype{z}[1]{>{\raggedleft\arraybackslash}p{#1pt}}

\title{DeCo: Frequency-Decoupled Pixel Diffusion for End-to-End Image Generation}

\author{
  \vspace{-20pt}\\
  \textbf{Zehong~Ma$^{1,3,\dag}$,\quad Longhui Wei$^{3,\ddagger,*}$,\quad Shuai Wang$^{2}$,\quad Shiliang Zhang$^{1,*}$,\quad Qi Tian$^{3}$} \vspace{3pt} \\
  $^1$ State Key Laboratory of Multimedia Information Processing, \\School of Computer Science, Peking University \quad $^2$Nanjing University \quad $^3$Huawei Inc.\vspace{3pt} \\
}

\begin{document}

\twocolumn[{%
\renewcommand\twocolumn[1][]{#1}%
\maketitle
\vspace{-28pt}
\begin{center}
    \centering
    \captionsetup{type=figure}
    \includegraphics[width=1\linewidth]{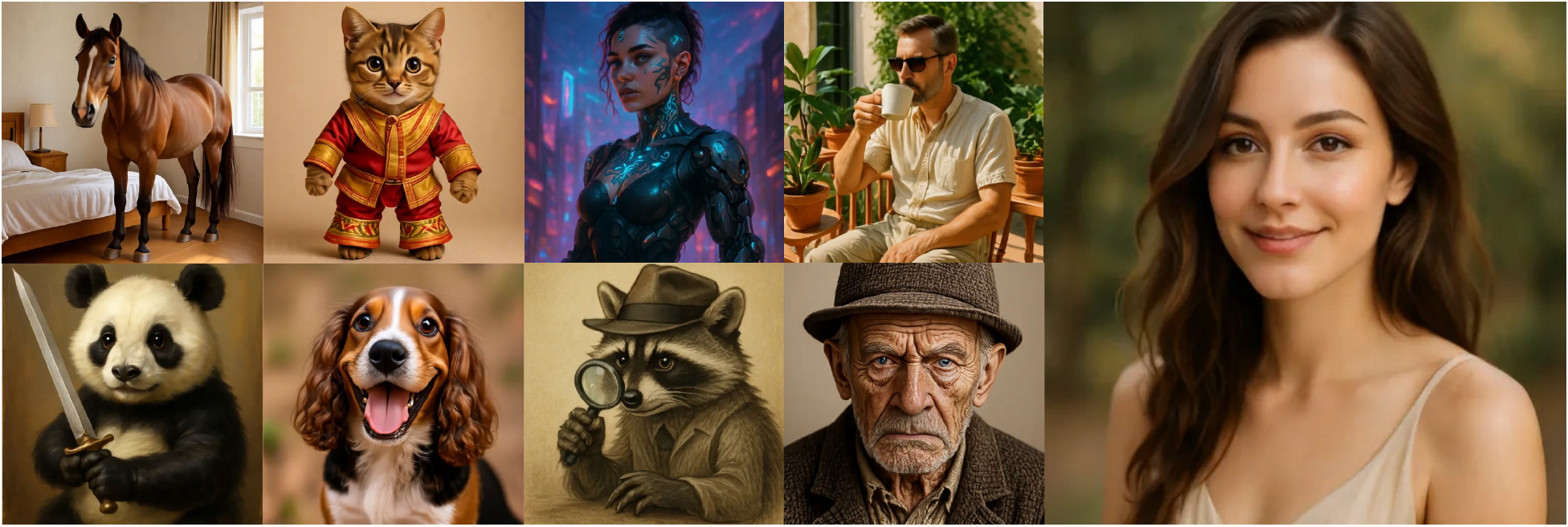}
    \vspace{-20pt}
    \captionof{figure}{Qualitative results of text-to-image generation of DeCo. All images are 512$\times$512 resolution.}
    \label{fig:t2i_visualization}
\end{center}%
}]

{
  \renewcommand{\thefootnote}{\fnsymbol{footnote}} 
  \footnotetext[1]{Corresponding authors. $\ddagger$ Project leader. $\dag$ Work was done during internship at Huawei Inc.}
}

\vspace{10pt}

\begin{abstract}
Pixel diffusion aims to generate images directly in pixel space in an end-to-end fashion. This approach avoids the limitations of VAE in the two-stage latent diffusion, offering higher model capacity. Existing pixel diffusion models suffer from slow training and inference, as they usually model both high-frequency signals and low-frequency semantics within a single diffusion transformer (DiT).
To pursue a more efficient pixel diffusion paradigm, we propose the frequency-\textbf{DeCo}upled pixel diffusion framework. With the intuition to decouple the generation of high and low frequency components, we leverage a lightweight pixel decoder to generate high-frequency details conditioned on semantic guidance from the DiT. This thus frees the DiT to specialize in modeling low-frequency semantics.
In addition, we introduce a frequency-aware flow-matching loss that emphasizes visually salient frequencies while suppressing insignificant ones.
Extensive experiments show that DeCo achieves superior performance among pixel diffusion models, attaining FID of 1.62 (256$\times$256) and 2.22 (512$\times$512) on ImageNet, closing the gap with latent diffusion methods. Furthermore, our pretrained text-to-image model achieves a leading overall score of 0.86 on GenEval in system-level comparison. Codes are publicly available at \url{https://github.com/Zehong-Ma/DeCo}.
\end{abstract}

\section{Introduction}
\label{sec:intro}

\begin{figure*}
    \centering
    \includegraphics[width=0.9\linewidth]{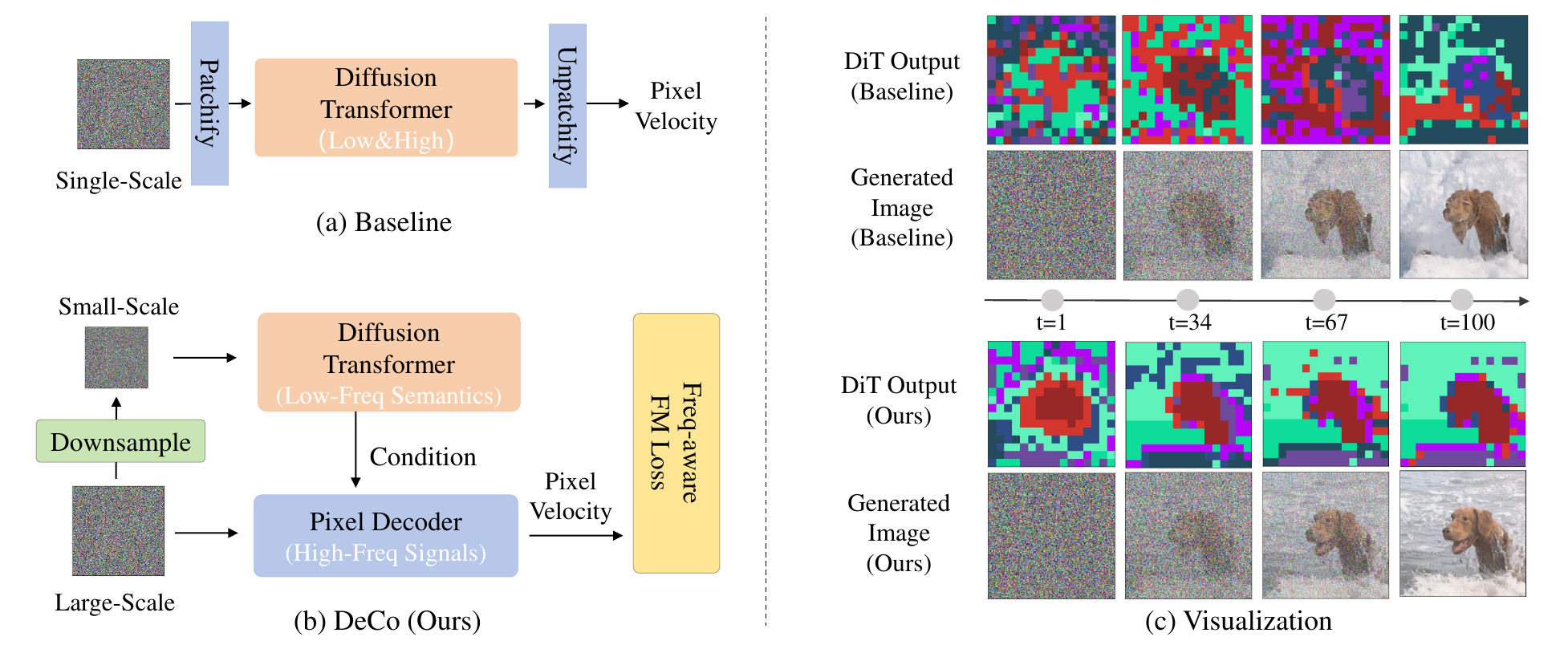}
    \vspace{-10pt}
    \caption{Illustration of our frequency-decoupled (DeCo) framework. In (a), traditional baseline models rely on a single DiT to jointly model both low-frequency semantics and high-frequency signals. (b) shows our DeCo framework, where a lightweight pixel decoder focuses on the high-frequency reconstruction, and the DiT models low-frequency semantics. As shown in (c), decoupling DiT from modeling high-frequency signals leads to better low-frequency semantic features in DiT Output, and higher image quality. 
    }
    \vspace{-10pt}
    \label{fig:intro}
\end{figure*}

Diffusion models~\cite{ddpm, ddim, wu2025customcrafter, wu2024spherediffusion, adm} have achieved remarkable success in high-fidelity image generation, offering exceptional quality and diversity. Research in this field generally follows two main directions: {latent diffusion} and {pixel diffusion}.
Latent diffusion models~\cite{ldm, dit, sit, flux2024, wu2024videomaker} split generation into two stages. A VAE first compresses images into a compact latent space, and a diffusion model operates within this space.
However, their performance is largely constrained by the VAE’s reconstruction quality and latent distribution~\cite{vavae, repa_e}. Training a VAE often requires adversarial or additional supervision~\cite{wang2025pixnerd}, which can be unstable and cause inevitable low-level artifacts.

Pixel diffusion avoids these VAE-dependent limitations by directly modeling raw pixels in an end-to-end manner. This enables more optimal distribution learning and eliminates artifacts from imperfect VAE compression.
However, it is challenging for pixel diffusion to jointly model complex high-frequency signals and low-frequency semantics within the enormous pixel space. As illustrated in ~\cref{fig:intro}~(a), traditional methods~\cite{adm, simple_diffusion, vdm, rdm, pixelflow} typically rely on a single model like diffusion transformer (DiT) to learn these two components from a single-scale input for each timestep.
The complex high-frequency signals, particularly high-frequency noise, could be hard to learn~\cite{wang2022fregan, skorokhodov2025improving, li2025jit}. They could also distract the DiT from learning low-frequency semantics~\cite{highfreq_gan, wang2020high, li2025jit}. 
As illustrated in \cref{fig:intro}~(c), this paradigm leads to noisy DiT outputs and degraded image quality.

Previous works~\cite{denton2015deep_multi_reso, karras2018progressive_multi_reso, pixelflow, rdm} demonstrate that it's more effective to reconstruct high-frequency signals from high-resolution input and model low-frequency semantics from low-resolution input. Other studies~\cite{park2022how, si2022inception, chenvision} show that transformers tend to capture low-frequency semantics well but struggle with high-frequency signals. We thus propose to decouple the generation of high and low frequency components. As illustrated in \cref{fig:intro}~(b), DeCo utilizes the DiT to specialize in low-frequency semantic modeling with downsampled inputs. Semantic cues are hence incorporated with a lightweight pixel decoder to reconstruct high-frequency signals. In other words, the pixel decoder takes the low-frequency semantics from DiT as condition and predicts pixel velocities with a high-resolution input. This new paradigm hence frees the DiT to specialize in modeling semantics, and allows for more specialized details generation. 

To further emphasize visually salient frequencies and suppress perceptually insignificant high-frequency components, we introduce a frequency-aware Flow-Matching (FM) loss inspired by JPEG~\cite{JPEG}. 
Unlike the standard FM loss, which treats all frequencies equally, 
our frequency-aware variant transforms the pixel velocity into the frequency domain using a discrete cosine transform and assigns adaptive weights to each frequency band. The adaptive weights are derived from JPEG quantization tables, which encode robust priors about the visual importance of different frequencies~\cite{JPEG}. By emphasizing visually salient frequencies and suppressing high-frequency noise, this loss simplifies the optimization landscape and enhances the visual fidelity.
\emph{It is worth noting that our motivation largely aligns with the concurrent work JiT}~\cite{li2025jit}.
JiT provides an explicit formula to decouple the high-frequency noise and clean image, while our DeCo provides a high-freq shortcut via the pixel decoder, enabling an implicit frequency decoupling. More discussions are included in Appendix~\ref{appendix: comparison_with_jit}.


We have conducted extensive experiments to test the performance of DeCo. It achieves superior results among pixel diffusion models, with FID scores of 1.62 (256×256) and 2.22 (512×512) on ImageNet, closing the gap with two-stage latent diffusion methods. Our pretrained text-to-image model also achieves leading results on GenEval (0.86) and DPG-Bench (81.4) in system-level evaluation. In summary, our contributions can be summarized as two aspects: i) we introduce a novel frequency-decoupled framework DeCo for pixel diffusion, where a lightweight pixel decoder is proposed to model high-frequency signals, freeing the DiT to specialize in low-frequency semantic modeling, and  ii) a novel frequency-aware FM loss is proposed to prioritize perceptually important frequencies, simplifying the training and improving visual quality. The strong performance verifies the effectiveness of decoupling the modeling of high and low frequency components in pixel diffusion.

\section{Related Work}

This work is closely related to latent diffusion, pixel diffusion, and frequency-decoupled image generation. 

\noindent\textbf{Latent Diffusion.}
Latent diffusion trains diffusion models in a compact latent space learned by a VAE~\cite{ldm}. Compared to raw pixel space, the latent space significantly reduces spatial dimensionality, easing learning difficulty and computational cost~\cite{ldm, vavae, dcae}. Consequently, VAEs have become a fundamental component in modern diffusion models~\cite{dit, sit, decoupled_dit, edm2, dod, flowdcn, wu2025multicrafter, dim, dmm, ma2025magcache, tian2025dic, wang2025mosa, wang2025mv}. However, training VAEs often involves adversarial objectives and perceptual supervision, which complicate the overall pipeline~\cite{wang2025pixnerd}. Poorly trained VAEs can produce decoding artifacts~\cite{sid, pixelflow}, limiting the generalization of latent diffusion models.
Early latent diffusion models mainly used U-Net-based architectures~\cite{tian2024udits, tian2025urepa}. The pioneering DiT~\cite{dit} introduced transformers into diffusion models, replacing the U-Net~\cite{uvit, adm}. 
SiT~\cite{sit} further validated the DiT with linear flow diffusion.
Subsequent works enhanced latent diffusion through representation alignment and joint optimization. REPA~\cite{repa} aligns intermediate features with a pretrained DINOv2~\cite{dinov2} model to learn better low-frequency semantics, which is compatible with our framework and is applied in both our baseline and final DeCo. REPA-E~\cite{repa_e} attempts to jointly optimize the VAE and DiT in an end-to-end fashion. However, this approach may suffer from training collapse with diffusion loss, as the continually changing latent space leads to unstable denoising targets. In contrast, pixel diffusion denoises in a fixed space, ensuring consistent targets and stable training.

\noindent\textbf{Pixel Diffusion.}
Pixel diffusion has progressed much more slowly than its latent counterparts due to the vast dimensionality of pixel space~\cite{adm, vdm, simple_diffusion}. Early pixel diffusion models typically rely on long residual connections~\cite{rdm, adm}, which may hinder scalability~\cite{wang2025pixnerd}. Recent attempts split the diffusion process into chunks at different resolution scales to reduce computational costs~\cite{pixelflow, rdm}.
Early methods split the diffusion process into multiple resolution stages~\cite{pixelflow, rdm}. Relay Diffusion~\cite{rdm} trains separate models for each scale, leading to higher cost and two-stage optimization. Pixelflow~\cite{pixelflow} uses one model across all scales and needs a complex denoising schedule that slows down inference.
Alternative approaches explore different model architectures. FractalGen~\cite{fractal} builds fractal generative models by recursively applying atomic modules, achieving self-similar pixel-level generation. TarFlow~\cite{tarflow} introduces a transformer-based normalizing flow to directly model and generate pixels. Recent PixNerd~\cite{wang2025pixnerd} employs a DiT to predict neural field parameters for each patch, rendering pixel velocities akin to test-time training. JiT~\cite{li2025jit} predicts the clean image to anchor generation to the low-dimensional data manifold. PixelGen~\cite{ma2026pixelgen} introduces perceptual supervision to simplify the training of pixel diffusion.

\noindent\textbf{Frequency-Decoupled Image Generation.}
Multi-scale cascaded methods~\cite{pixelflow, rdm} can be approximately regarded as a form of temporal frequency decoupling, i.e., early steps generate low-frequency semantics, and later steps refine high-frequency details. 
However, these methods still use a single model or architecture to learn all frequencies for each timestep and the high-frequency signals still exist.
High-frequency noise may interfere with low-frequency semantic learning. They also rely on complex denoising schedules and small patch sizes, which reduce training or sampling efficiency.
Our DeCo introduces an explicit frequency decoupling in architecture. Instead of separating distinct frequencies across timesteps, DeCo simultaneously processes them within each timestep in an end-to-end manner.
Recent two-stage work DDT~\cite{decoupled_dit} explores single-scale frequency decoupling in a compressed latent space, showing that frequency decoupling remains important even in compressed space. Unlike DDT, our DeCo is a multi-scale design for pixel diffusion. Our decoder uses attention-free linear layers instead of DDT’s attention-based DiT blocks, making it more efficient in large-scale inputs.

\section{Method}
\begin{figure*}
    \centering
    \includegraphics[width=0.9\linewidth]{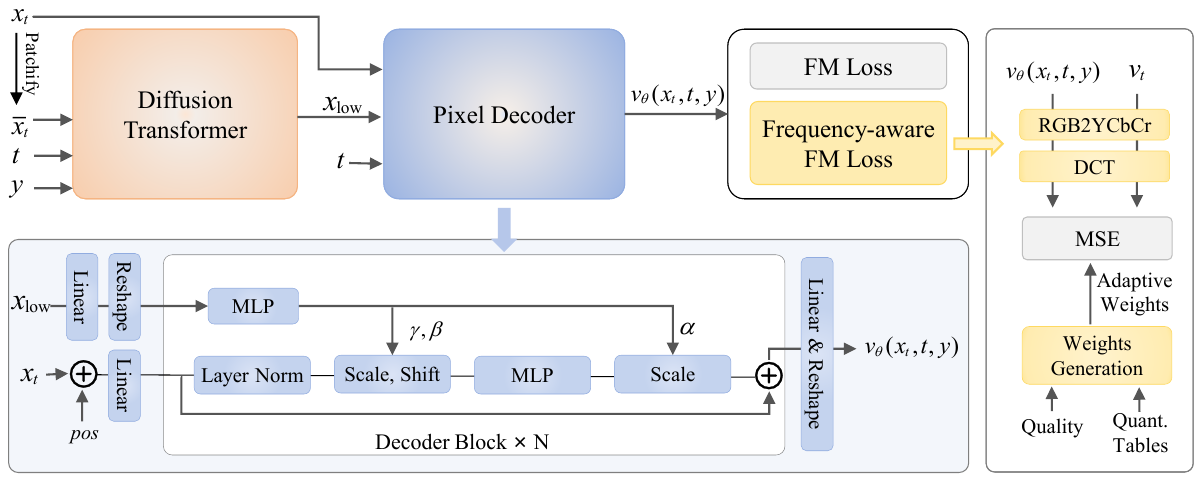}
    \vspace{-5pt}
    \caption{Overview of the proposed frequency-decoupled (DeCo) framework. The DiT operates on downsampled inputs to model low-frequency semantics, while the lightweight pixel decoder generates high-frequency details under the DiT's semantic guidance.}
\label{fig:architecture}
\vspace{-10pt}
\end{figure*}

\subsection{Overview}
\label{sec:method_overview}
This part first reviews the conditional flow matching in baseline pixel diffusion, then proceeds to introduce our frequency-decoupled pixel diffusion framework.

\noindent\textbf{Conditional Flow Matching.}
The conditional flow matching~\citep{lipman2023flow, luo2024latent} provides a continuous-time generative modeling framework that learns a velocity field ${v}_\theta({x}, t, y)$ to transport samples from a simple prior distribution (e.g., Gaussian) to a data distribution conditioned on the label ${y}$ and time $t$.
Given a forward trajectory ${x}_t$ by an interpolation between a clean image ${x}_0$ and noise ${x}_1$, the objective of conditional flow matching is to match the model-predicted velocity ${v}_\theta({x}_t, t, {y})$ to the ground-truth velocity ${v}_t$:

\begin{equation}
\mathcal{L}_{\mathrm{FM}} = \mathbb{E}_{{x}_t, t, {y}} \left[ \left\| {v}_\theta({x}_t, t, {y}) - {v}_t \right\|^2 \right],
\end{equation}
where the linear interpolation of trajectory  ${x}_t$ is defined as: 
\begin{equation}
{x}_t = (1 - t)\,{x}_0 + t\,{x}_1.
\end{equation}
The ground-truth velocity ${v}_t$ can be derived from ${x}'_t$, i.e., the time derivative of ${x}_t$:
\begin{equation}
{v}_t = {x}'_t = {x}_1-{x}_0.
\end{equation}

In the pixel diffusion baseline, the trajectory ${x}_t$ is usually first patchified into tokens by a patch embedding layer~\cite{pixelflow, dit} instead of a VAE to downsample the image. 
In our baseline and DeCo experiments, we use the same patch size of 16 for the DiT's input. 
In baseline, the patchified trajectory ${\bar{x}}_t$ is then fed into the DiT to predict the pixel velocity with an unpatchify layer. The DiT is required to simultaneously model both the high-frequency signals and low-frequency semantics. The high-frequency signals, particularly the high-frequency noise, are hard to model, which can distract the DiT from learning low-frequency semantics.

\noindent\textbf{DeCo.} To separate high-frequency generation from low-frequency semantic modeling, we propose a frequency-decoupled framework DeCo.
As illustrated in \cref{fig:architecture}, the DiT is utilized to generate low-frequency semantics $c$ from downsampled low-resolution inputs $\bar{x}_t$ as follows: 
\begin{equation}
    {x_\text{low}} = \text{DiT}({\bar{x}}_t, t, y),
\end{equation}
where $t$ is time and $y$ is the label or textual prompt. As depicted in \cref{sec:method_pixel_decoder}, a lightweight pixel decoder then takes the low-frequency semantics $c$ from DiT as a condition to generate additional high-frequency details with a full-resolution dense input $x_t$, predicting the final pixel velocity as follows:
\begin{equation}
    {v}_\theta({x}_t, t, y) = \text{Dec}({x}_t, t, {x_\text{low}}).
\end{equation}

This new paradigm leverages the pixel decoder to generate high-frequency details, freeing the DiT to specialize in modeling semantics. The decoupling disentangles the modeling of different frequencies into different modules, leading to faster training and improved visual fidelity.



To further emphasize visually salient frequencies and ignore insignificant high-frequency components, we introduce a frequency-aware Flow-Matching (FM) Loss $\mathcal{L}_{\mathrm{FreqFM}}$ as depicted in \cref{sec: method_freqfm_loss}. This loss reweights different frequency components with the adaptive weights derived from JPEG perceptual priors~\cite{JPEG}.
Combined with the standard pixel-level flow-matching loss and the REPA~\cite{repa} alignment loss from the baseline, the final objective can be represented as:

\begin{equation}
\label{eq: overall_loss}
\mathcal{L} = \mathcal{L}_{\mathrm{FM}} + \mathcal{L}_{\mathrm{FreqFM}} + \mathcal{L}_{\mathrm{REPA}}
\end{equation}

\subsection{Pixel Decoder}
\label{sec:method_pixel_decoder}
As illustrated in \cref{fig:architecture}, the pixel decoder is a lightweight attention-free network composed of $N$ linear decoder blocks and several linear projection layers.
All operations are local and linear, enabling efficient high-frequency modeling without the computational overhead of self-attention.

\noindent\textbf{Dense Query Construction.}
The pixel decoder directly takes the full-resolution noised image as input, without downsampling. All noised pixels are concatenated with their corresponding positional embeddings $pos$ and linearly projected by $W_{\text{in}}$ to form dense query vectors $h_0$: 
\begin{equation}
    h_0 = W_{\text{in}}(\text{Concat}({x}_t, pos)),
\end{equation}
where $h_0 \in \mathbb{R}^{B \times H \times W \times d}$, with ${H}$ and ${W}$ denoting the original image height and width (e.g., 256), and ${d}$ representing the hidden dimension of pixel decoder (e.g., 32). See~\cref{exp: tab_ablation}~(c) and (d) for related ablation studies.

\noindent\textbf{Decoder Block.}
For each decoder block, the DiT output ${x_\text{low}}$ is linearly upsampled and reshaped to match the spatial resolution of ${x}_t$, yielding ${x_\text{low}^{\text{up}}}$.
A MLP then generates modulation parameters $\alpha, \beta, \gamma$ for AdaLN:

\begin{equation}
    \alpha, \beta, \gamma = \text{MLP}(\sigma(x_\text{low}^{\text{up}}+t)),
\end{equation}
where the $\sigma$ is the SiLU activation function.
We utilize the AdaLN-Zero~\cite{dit} to modulate the dense decoder queries in each block as follows:

\begin{equation}
h_{\text{N}} = h_{\text{N-1}}+\alpha*(\text{MLP}((1+\gamma)*h_{\text{N-1}} + \beta)),
\end{equation}
where the MLP contains two linear layers with SiLU~\cite{elfwing2018sigmoid}.

\noindent\textbf{Velocity Prediction.}
Finally, a linear projection followed by a rearrangement operation maps the decoder output to the pixel space, yielding the predicted velocity $\mathbf{v}_\theta(\mathbf{x}_t, t, y)$. The velocity encompasses the high-frequency details generated by the pixel decoder and the semantic cues from DiT.


\subsection{Frequency-aware FM Loss}
\label{sec: method_freqfm_loss}
To further encourage the pixel decoder to focus on perceptually important frequencies and suppress insignificant noise, we introduce a frequency-aware flow-matching (FM) loss.

\noindent\textbf{Spatial–Frequency Transformation.} 
We first transform both the predicted and ground-truth pixel velocities from the spatial domain to the frequency domain. This is done by converting the color space to YCbCr and applying a block-wise $8 \times 8$ discrete cosine transform (DCT), following JPEG~\cite{JPEG}. Denoting this transformation as $\mathcal{T}$, we have:

\begin{equation}
\mathbb{V}_\theta = \mathcal{T}({v}_\theta({x}_t, t, y)),
\quad
\mathbb{V}_t = \mathcal{T}({v}_t).
\end{equation}



\noindent\textbf{Perceptual Weighting.}
To emphasize visually salient frequencies while suppressing insignificant ones, we employ the JPEG quantization tables~\cite{JPEG} as visual priors to generate adaptive weights. Frequencies with smaller quantization intervals are more perceptually important. Thus, we use the normalized reciprocal of the scaled quantization tables $Q_{\text{cur}}$ in quality $q$ as adaptive weights $w$, i.e., $w=\frac{1}{Q_{\text{cur}}}/\mathbb{E}(\frac{1}{Q_\text{cur}})$.
When the quality $q$ is between 50 and 100, the scaled quantization tables $Q_{\text{cur}}$ in quality $q$ can be acquired following JPEG’s predefined rules~\cite{JPEG}:
\begin{equation}
    Q_{\text{cur}} = \max\left(1, \left\lfloor \frac{Q_{\text{base}} \cdot (100-q) + 25}{50} \right\rfloor\right),
    \label{eq:quant_table}
\end{equation}
where $Q_{\text{base}}$ denotes the standard base quantization tables defined in the JPEG specification~\cite{JPEG}.
With the adaptive weights $w$, the frequency-aware FM loss is defined as:
\begin{equation}
\mathcal{L}_{\mathrm{FreqFM}} = \mathbb{E}_{{x}_t, t, {y}} \left[ w \left\| \mathbb{V}_\theta - \mathbb{V}_t \right\|^2 \right]
\end{equation}

\subsection{Empirical Analysis}

\begin{figure}
    \centering
    \includegraphics[width=0.95\linewidth]{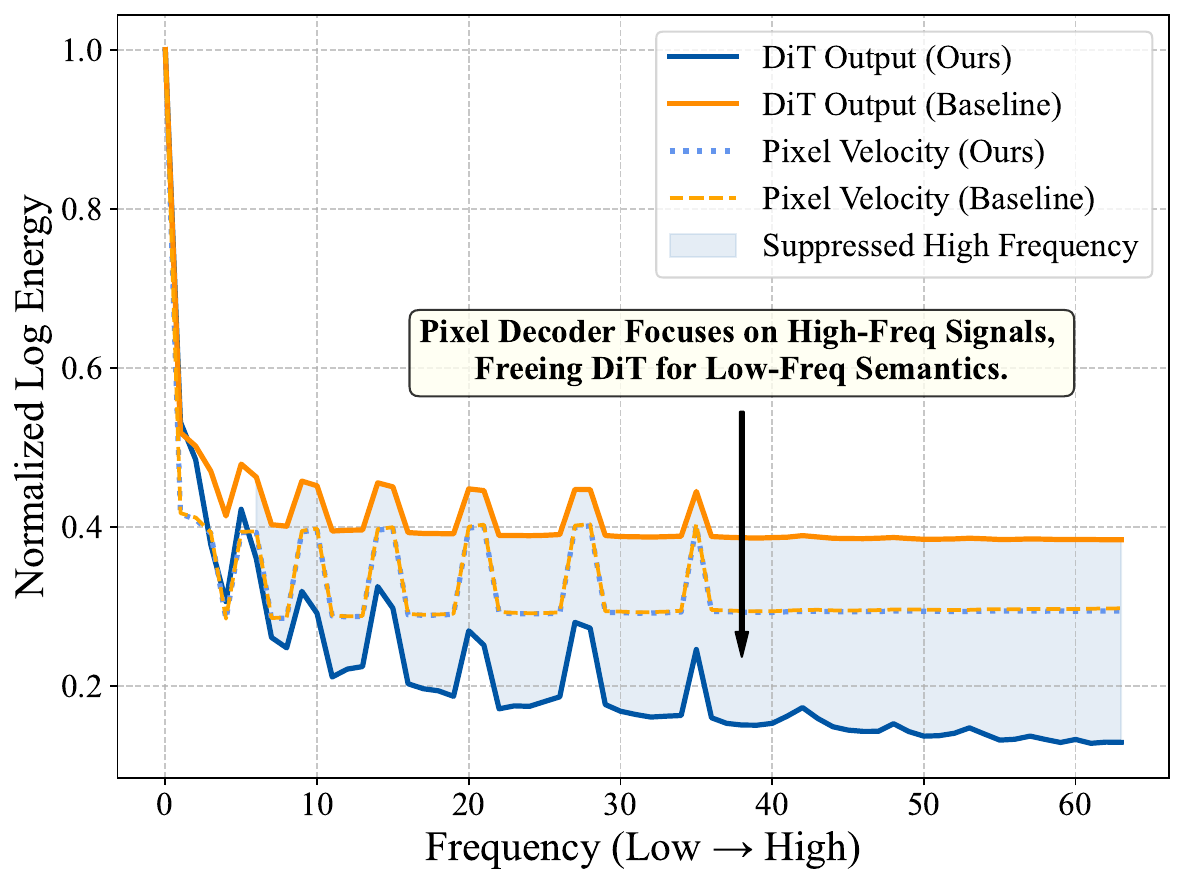}
    \vspace{-5pt}
    \caption{DCT energy distribution of DiT outputs and predicted pixel velocities. Compared with baseline, DeCo suppresses high-frequency signals in DiT outputs while preserving strong high-frequency energy in pixel velocity, confirming effective frequency decoupling. The distribution is computed on 10K images across all diffusion steps using DCT transform with 8$\times$8 block size.
    }
    \vspace{-10pt}
    \label{fig:frequency_plot}
\end{figure}

To verify that DeCo effectively decouples frequencies, we analyze the DCT energy spectra of the DiT outputs and the pixel velocity, as shown in \cref{fig:frequency_plot}. Compared to the baseline, our pixel decoder successfully maintains all frequency components in pixel velocity. Meanwhile, the DiT outputs in DeCo exhibit significantly lower high-frequency energy than those of the baseline, indicating high-frequency components have been shifted away from the DiT and into the pixel decoder. These observations confirm that DeCo performs effective frequency decoupling.
Results from \cref{exp: tab_ablation}~(c) and (d) further show that this successful decoupling benefits from two key architectural designs.

\noindent\textbf{Multi-scale Input Strategy.} The multi-scale input strategy is critically important. With this strategy, the pixel decoder can easily model high-frequency signals on high-resolution original inputs, freeing the DiT modeling the low-frequency semantics from low-resolution inputs where high-frequency signals have been partly suppressed. 

\noindent\textbf{AdaLN-based Interaction.}
AdaLN provides a powerful interaction mechanism between DiT and pixel decoder.
In our framework, the DiT provides stable, low-frequency semantic conditioning. The AdaLN layer then uses the DiT output as condition to modulate the dense query features in the pixel decoder. Our experiments confirm that this modulation is more effective than the simpler method, such as upsampling and adding the low-frequency features to their high-frequency counterparts like UNet.
\section{Experiments}
\label{sec:experiments}

\begin{table*}[t]
\centering
\caption{Comparison with the baseline and other recent methods. \textcolor{gray}{Text in gray}: latent diffusion models that require VAE. $\dagger$: use 100 steps.}
\label{tab:baseline_comparison}
\small
\setlength{\tabcolsep}{5pt}
\begin{tabular}{l|c|cc|ccc|cccc}  
\hline
 & & \multicolumn{2}{c|}{Train} & \multicolumn{3}{c|}{Inference} & \multicolumn{4}{c}{Generation Metrics} \\   
Method & Params & Mem (GB) & Speed (s/it) & \begin{tabular}[c]{@{}c@{}}Mem (GB)\end{tabular} & \begin{tabular}[c]{@{}c@{}}1 image\end{tabular} & \begin{tabular}[c]{@{}c@{}}1 iter\end{tabular} & FID$\downarrow$ & sFID$\downarrow$ & IS$\uparrow$  & Rec.l$\uparrow$ \\ \hline 
\textcolor{gray}{DiT-L/2}~\cite{dit} & \textcolor{gray}{458M+ 86M} & \textcolor{gray}{28.5} & \textcolor{gray}{0.43} & \textcolor{gray}{3.5} & \textcolor{gray}{0.63s} & \textcolor{gray}{0.013s} & \textcolor{gray}{41.93} & \textcolor{gray}{13.76} & \textcolor{gray}{36.52} & \textcolor{gray}{0.59} \\
PixDDT~\cite{decoupled_dit} & 434M & 23.6 & 0.22 & 2.4  &  0.49s &  0.010s & 46.37  &  17.14  &  36.24  & 0.63 \\
PixNerd~\cite{wang2025pixnerd} & 458M & 29.2 & 0.25 & 2.6 & 0.48s & 0.010s & 37.49 & 10.65 & 43.01 & 0.62 \\
PixelFlow~\cite{pixelflow} & 459M & 73.2 & 1.61 & 3.9 & 6.61s$\dagger$ & 0.066s & 54.33 & 9.71 & 24.67 & 0.58 \\ 
JiT+REPA~\cite{li2025jit} & 459M & 24.8 & 0.23 & 2.5 & 0.46s & 0.009s & 39.06 & 11.45 & 39.57 & 0.63 \\ 
\hline  
Baseline & 459M & 24.8 & 0.22 & 2.5 & 0.46s & 0.009s & 61.10 & 15.86 & 16.81 & 0.60 \\  
DeCo w/o REPA & 426M  & 26.4 & 0.22 & 2.4 & 0.46s & 0.009s & 67.55 & 10.58 & 19.10 & 0.56\\
DeCo w/o $\mathcal{L}_\mathrm{FreqFM}$ & 426M & 27.5 & 0.24 & 2.4 & 0.46s & 0.009s & 34.12 & 10.41 & 46.44 & 0.64 \\  
DeCo & 426M & 27.5 & 0.24 & \textbf{2.4} & \textbf{0.46s} & \textbf{0.009s} & \textbf{31.35} & \textbf{9.34} & \textbf{48.35} & \textbf{0.65} \\ \hline  
\end{tabular}
\vspace{-5pt}
\end{table*}


We conduct ablation studies and baseline comparisons on ImageNet $256\times256$. For class-to-image generation, we provide detailed comparisons on ImageNet $256\times256$ and $512\times512$, and report FID~\cite{fid}, sFID~\cite{sfid}, IS~\cite{is}, Precision, and Recall~\cite{pr_recall}. For text-to-image generation, we report results on GenEval~\cite{geneval} and DPG-Bench~\cite{dpg}.

\begin{figure}[t]
    \centering
    \includegraphics[width=0.9\linewidth]{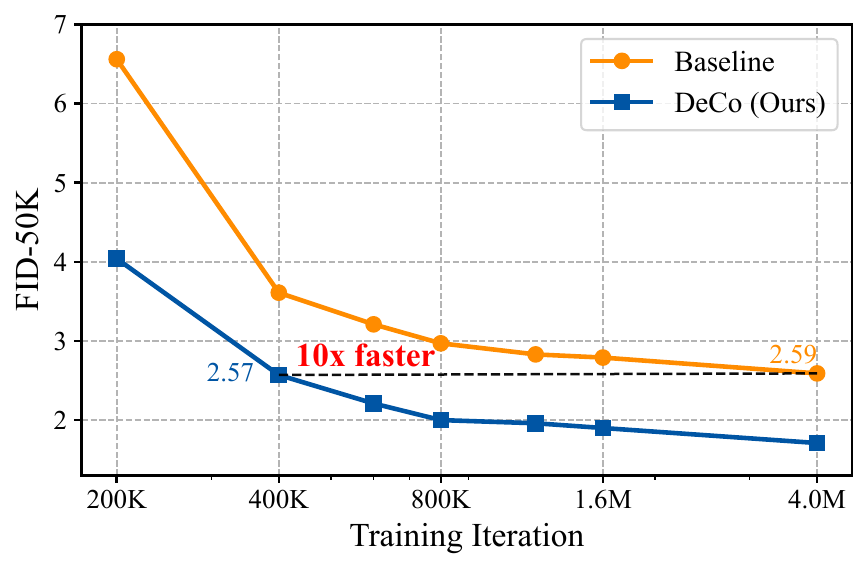}
    \vspace{-10pt}
    \caption{FID comparison between our DeCo and baseline. DeCo reaches 2.57 FID in 400k iterations, 10× faster than the baseline.}
    \label{fig:fid_comparison}
    \vspace{-15pt}
\end{figure}

\begin{figure*}[t]
    \centering
    \includegraphics[width=0.99\linewidth]{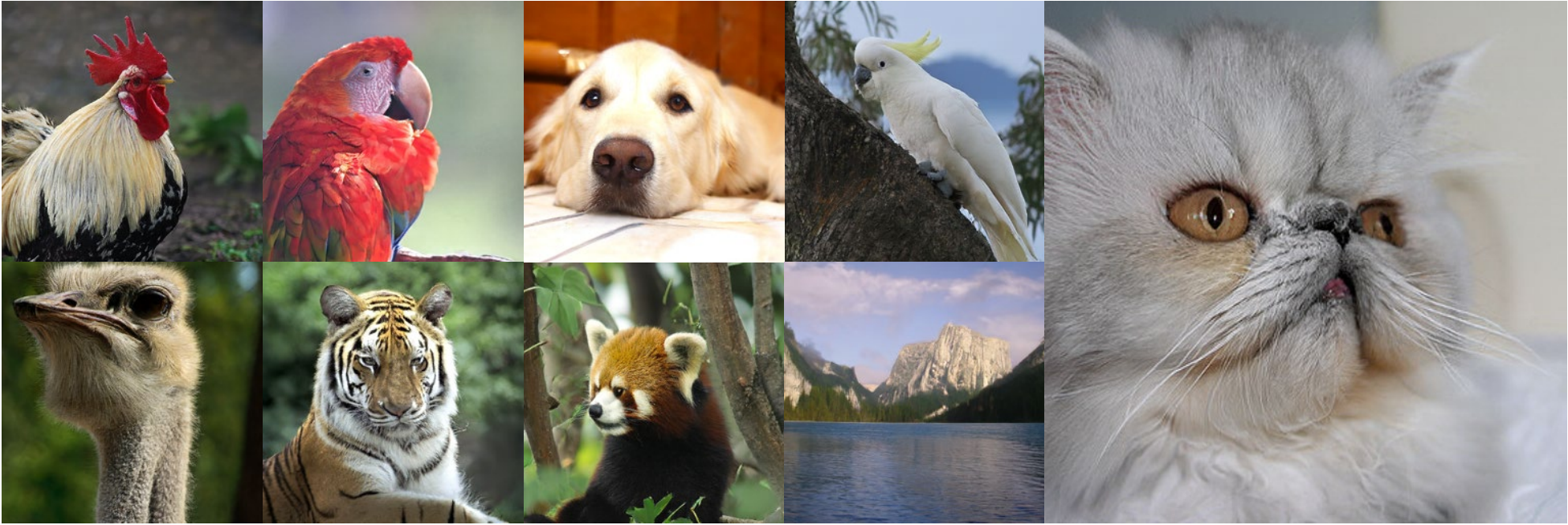}
    \vspace{-5pt}
    \caption{Qualitative results of class-to-image generation of DeCo. All images are 256$\times$256 resolution.}
    \label{fig:c2i_visualization}
    \vspace{-10pt}
\end{figure*}

\subsection{Comparison with Baselines}
\label{sec:exp_baseline}
\noindent\textbf{Setup.} In the baseline comparisons, all diffusion models are trained on ImageNet at a 256×256 resolution for 200k iterations, using a large DiT variant.
The key architectural modification from the baseline is the replacement of the final two DiT blocks with our proposed pixel decoder.
For inference, we use 50 Euler steps without classifier-free guidance~\cite{cfg} (CFG).
We compare against the two-stage DiT-L/2~\cite{dit} that requires a VAE, and other recent pixel diffusion models such as PixelFlow~\cite{pixelflow} and PixNerd~\cite{wang2025pixnerd}. We also adapt DDT~\cite{decoupled_dit} into pixel diffusion to create a PixDDT baseline. Besides, we intergrate recent JiT~\cite{li2025jit} in our baseline with REPA~\cite{repa} for a fair comparison.
Please refer to Appendix \ref{appendix: baseline_train_details} for more details.


\begin{table*}[t] \centering 
\vspace{-5pt}
\small \caption{Class-to-image generation on ImageNet 256$\times$256 and 512$\times$512 with CFG. DeCo achieves superior performance in end-to-end pixel diffusion and is competitive with two-stage latent diffusion models. \textcolor{gray}{Text in gray}: latent diffusion models that require VAE. \textcolor{blue!40}{Text in blue background}: inference with Heun sampler~\cite{heun1900neue} and 50 sampling steps.
} 
\renewcommand{\arraystretch}{1.1} 
\begin{tabular}{c|l|c c c c| c c c c c}
\toprule
\multicolumn{1}{c}{} & & \textbf{Params} & \textbf{Epochs} & \textbf{NFE} & \textbf{Latency(s)} & \textbf{FID}$\downarrow$ & \textbf{sFID}$\downarrow$ & \textbf{IS}$\uparrow$ & \textbf{Pre.}$\uparrow$ & \textbf{Rec.}$\uparrow$ \\
\midrule
\multirow{16}{*}{\rotatebox[origin=c]{90}{{256$\times$256}}} & \textcolor{gray}{DiT-XL/2}~\cite{dit} & \textcolor{gray}{675M + 86M} & \textcolor{gray}{1400} & \textcolor{gray}{250$\times$2} & \textcolor{gray}{3.44} & \textcolor{gray}{2.27} & \textcolor{gray}{4.60} & \textcolor{gray}{278.2} & \textcolor{gray}{0.83} & \textcolor{gray}{0.57} \\
& \textcolor{gray}{SiT-XL/2}~\cite{sit} & \textcolor{gray}{675M + 86M} & \textcolor{gray}{1400} & \textcolor{gray}{250$\times$2} & \textcolor{gray}{3.44} & \textcolor{gray}{2.06} & \textcolor{gray}{4.50} & \textcolor{gray}{284.0} & \textcolor{gray}{0.83} & \textcolor{gray}{0.59} \\
& \textcolor{gray}{REPA-XL/2}~\cite{repa} & \textcolor{gray}{675M + 86M} & \textcolor{gray}{800} & \textcolor{gray}{250$\times$2} & \textcolor{gray}{3.44} & \textcolor{gray}{1.42} & \textcolor{gray}{4.70} & \textcolor{gray}{305.7} & \textcolor{gray}{0.80} & \textcolor{gray}{0.64} \\
\arrayrulecolor{gray} \cline{2-11} \arrayrulecolor{black} %
& ADM~\cite{adm} & 554M & 400 & 250 & 15.2 & 4.59 & 5.25 & 186.7 & \textbf{0.82} & 0.52 \\
& RDM~\cite{rdm} & 553M + 553M & 400 & 250 & 38.4 & 1.99 & \textbf{3.99} & 260.4 & 0.81 & 0.58 \\
& JetFormer~\cite{jetformer} & 2.8B & - & - & - & 6.64 & - & - & 0.69 & 0.56 \\
& FractalMAR-H~\cite{fractal} & 848M & 600 & - & 155 & 6.15 & - & \textbf{348.9} & 0.81 & 0.46 \\
& PixelFlow-XL/4~\cite{pixelflow} & 677M & 320 & 120$\times$2 & 9.78 & 1.98 & 5.83 & 282.1 & 0.81 & 0.60 \\
& {PixNerd-XL/16}~\cite{wang2025pixnerd} & {700M} & {320} & {100$\times$2} & 1.18 & {1.95} & {4.54} & {300} & {0.80} & {0.60} \\
& Baseline-XL/16 & 700M & 320 & 100$\times$2 & 1.03 & 2.79 & 4.90 & 296.0 & 0.79 & 0.60 \\
& \textbf{DeCo-XL/16} & {682M} & {320} & {100$\times$2} & 1.05 & {1.90} & {4.47} & {303} & {0.80} & {0.61} \\
& \textbf{DeCo-XL/16} & {682M} & {800} & {100$\times$2} & 1.05 & {1.71} & {4.54} & {304} & {0.80} & {0.61} \\
& \textbf{DeCo-XL/16} & {682M} & {800} & {250$\times$2} & 2.63 & \textbf{1.62} & {4.41} & {301} & {0.80} & \textbf{0.62} \\ 

\rowcolor{blue!10} \cellcolor{white} & JiT-H/16 (Heun)~\cite{li2025jit} & 953M &  600 &  100$\times$2 & - & 1.86  & - & 303 & - & - \\
\rowcolor{blue!10} \cellcolor{white} & \textbf{DeCo-XL/16 (Heun)}  & {682M} & {600} & {100$\times$2} & 1.05 & {1.69} & {4.59} & {304} & {0.79} & {0.63} \\

\midrule
\multirow{8}{*}{\rotatebox[origin=c]{90}{{512$\times$512}}} %
& \textcolor{gray}{DiT-XL/2}~\cite{dit} & \textcolor{gray}{675M + 86M} & \textcolor{gray}{600} & \textcolor{gray}{250$\times$2} & \textcolor{gray}{11.1} & \textcolor{gray}{3.04} & \textcolor{gray}{5.02} & \textcolor{gray}{240.8} & \textcolor{gray}{0.84} & \textcolor{gray}{0.54} \\
& \textcolor{gray}{SiT-XL/2}~\cite{sit} & \textcolor{gray}{675M + 86M} & \textcolor{gray}{600} & \textcolor{gray}{250$\times$2} & \textcolor{gray}{11.1} & \textcolor{gray}{2.62} & \textcolor{gray}{4.18} & \textcolor{gray}{252.2} & \textcolor{gray}{0.84} & \textcolor{gray}{0.57} \\
\arrayrulecolor{gray} \cline{2-11} \arrayrulecolor{black} %
& ADM-G~\cite{adm} & 554M & 400 & 250 & 21.2 & 7.72 & 6.57 & 172.7 & \textbf{0.87} & 0.53 \\
& RIN~\cite{rin} & 320M & - & 250 & - & 3.95 &- & 210.0 & - & - \\
& SimpleDiffusion~\cite{sid} & 2B & 800 & 250×2 & - & 3.54 & - & 205.0 & - & - \\
& VDM++~\cite{vdm} & 2B & 800 & 250$\times$2 & - & 2.65 & - & 278.1 & -& - \\
& {PixNerd-XL/16}~\cite{wang2025pixnerd} & {700M} & {340} & {100$\times$2} & 2.47 & {2.84} & {5.95} & {245.6} & {0.80} & {0.59} \\
& \textbf{DeCo-XL/16} & 682M & 340 & 100$\times$2 & 2.25 & \textbf{2.22} &  \textbf{4.67} & \textbf{290.0} & 0.80 & \textbf{0.60} \\
\bottomrule
\end{tabular}
\vspace{-5pt}
\end{table*}

\begin{table*}[ht]
\centering
\caption{Text-to-image generation on GenEval~\cite{geneval} and DPG-Bench~\cite{dpg} at a 512$\times$512 resolution.}
\label{tab:my-table}
\small
\setlength{\tabcolsep}{5pt}
\begin{tabular}{l|c|ccccccc|c}
\toprule
 & Diffusion & \multicolumn{7}{c|}{GenEval} & \multicolumn{1}{c}{DPG-Bench}  \\   
Method     & Params        & Sin.Obj. & Two.Obj & Counting & Colors & Pos & Color.Attr. & Overall$\uparrow$ & Average\\ \midrule
PixArt-$\alpha$~\cite{chen2023pixartalpha} & 0.6B & 0.98 & 0.50  & 0.44  & 0.80 &  0.08 &  0.07 & 0.48 & 71.1 \\
SD3~\cite{sd3}  & 8B        & 0.98                & 0.84             & 0.66           & 0.74         & 0.40           & 0.43                    & 0.68    & -     \\
FLUX.1-dev~\cite{flux2024} &  12B  & 0.99                & 0.81             & \textbf{0.79}           & 0.74         & 0.20           & 0.47                    & 0.67  & 84.0        \\
DALL-E 3~\cite{Betker2023ImprovingIG} & - & 0.96 & 0.87 &  0.47   &   0.83 &  0.43&  0.45  & 0.67 & 83.5\\
BLIP3o~\cite{blip3o}   & 4B         & -         & -                & -              & -            & -              & -                       & 0.81   &  79.4       \\
OmniGen2~\cite{wu2025omnigen2}      &   4B  & \textbf{1}                   & \textbf{0.95}             & 0.64           & 0.88         & 0.55           & 0.76                    & 0.80      &  83.6\\ 
\hline
PixelFlow-XL/4~\cite{pixelflow} &  882M &      -        & -                & -              & -            & -              & -                       & 0.60 & -     \\ 
PixNerd-XXL/16~\cite{wang2025pixnerd} &  1.2B &      0.97        & 0.86              & 0.44         & 0.83            & 0.71             & 0.53                      & 0.73    & 80.9  \\

\textbf{DeCo-XXL/16} & 1.1B & \textbf{1} & 0.92  & 0.72  & \textbf{0.91} & \textbf{0.80}  &  \textbf{0.79}   &  \textbf{0.86}  & 81.4 \\
        
        \bottomrule
\end{tabular}%

\end{table*}

\renewcommand{\thesubtable}{{\alph{subtable}}}
\begin{table*}[t]
\setlength{\tabcolsep}{5pt}
\centering
\caption{Ablation experiments on architecture design and hyper-parameters.}
\label{exp: tab_ablation}
\footnotesize
\subfloat[{Hidden Size $d$ of Pixel Decoder.}]
{
\begin{minipage}{0.32\linewidth}
\centering
\begin{tabular}{c|ccc}
\toprule
 Channel &FID$\downarrow$ & sFID & IS$\uparrow$ \\
\hline
16 & 37.63 & 10.64 & 41.54 \\
32 &  \textbf{34.12} & \textbf{10.41} & \textbf{46.44} \\
64 & 35.88& 10.79 & 44.07 \\
\bottomrule
\end{tabular}
\label{exp:ablation:channel}
\end{minipage}
}
\hspace{\fill}
\subfloat[{Depth $N$ of Pixel Decoder}.]{
\begin{minipage}{0.32\linewidth}
\centering
\begin{tabular}{c|ccc}
\toprule
Depth &FID$\downarrow$ &sFID$\downarrow$  & IS$\uparrow$ \\
\hline
1& 37.10 & 10.73 & 41.06 \\
3& \textbf{34.12} & \textbf{10.41} & \textbf{46.44} \\
6& 35.46&  10.82&44.60 \\
\bottomrule
\end{tabular}
\label{exp:ablation:depth}
\end{minipage}
}
\hspace{\fill}
\subfloat[Patch Size of Pixel Decoder]{
\begin{minipage}{0.32\linewidth}
\centering
\begin{tabular}{c|ccc}
\toprule
Patch Size &FID$\downarrow$ &sFID$\downarrow$ & IS$\uparrow$ \\
\hline
1& \textbf{31.35} & \textbf{9.34} & \textbf{48.35} \\
4& 34.39& 11.15& 45.53 \\
16& 55.59 & 44.16 & 34.44 \\
\bottomrule
\end{tabular}
\label{exp:ablation:dec_patch_size}
\end{minipage}
}
\hspace{\fill}

\vspace{5pt}
\subfloat[Interaction of DiT and Pixel Decoder.]{
\begin{minipage}{0.32\linewidth}
\centering
\begin{tabular}{c|ccc}
\toprule
Interaction &FID$\downarrow$&sFID$\downarrow$  & IS$\uparrow$ \\
\hline
AdaLN& \textbf{31.35} & \textbf{9.34} & \textbf{48.35} \\
Add& 36.02&  9.99& 41.74 \\
\bottomrule
\end{tabular}
\label{exp:ablation:adaln}
\end{minipage}
}
\hspace{\fill}
\subfloat[{Loss Weight of $\mathcal{L}_\mathrm{FreqFM}$}]{
\begin{minipage}{0.32\linewidth}
\centering
\begin{tabular}{c|ccc}
\toprule
Weight &FID$\downarrow$ &sFID$\downarrow$ & IS$\uparrow$ \\
\hline
0.5& 33.54& 10.27& 46.38 \\
1& \textbf{31.35} & \textbf{9.34}& \textbf{48.35} \\
2& 32.97& 9.42&46.55 \\
\bottomrule
\end{tabular}
\label{exp:ablation:loss_weight}
\end{minipage}
}
\hspace{\fill}
\subfloat[JPEG Quality in $\mathcal{L}_\mathrm{FreqFM}$.]{
\begin{minipage}{0.32\linewidth}
\centering
\begin{tabular}{c|ccc}
\toprule
Quality &FID$\downarrow$ &sFID$\downarrow$ & IS$\uparrow$ \\
\hline
50& 31.54 & {9.45} &47.70 \\
85& \textbf{31.35}& \textbf{9.34}&  \textbf{48.35} \\
100& 33.84& 10.74 & 46.14 \\
\bottomrule
\end{tabular}
\label{exp:ablation:jpeg_quality}
\end{minipage}
}
\vspace{-5pt}
\end{table*}

\noindent\textbf{Detailed Comparisons.} 
As detailed in \cref{tab:baseline_comparison}, our DeCo framework, despite having fewer parameters, significantly outperforms the baseline across all metrics while maintaining comparable training and inference costs. Notably, with the frequency-decoupled architecture alone, DeCo w/o $\mathcal{L}_\mathrm{FreqFM}$ lowers FID from 61.10 to 34.12 and raises the IS from 16.81 to 46.44 compared to the baseline. By additionally incorporating a frequency-aware FM loss, our DeCo further reduces the FID to 31.35 and achieves consistent gains on other metrics.
Compared to the two-stage DiT-L/2, our VAE-free DeCo model demonstrates substantially lower training and inference overhead while achieving comparable performance.
Against other pixel diffusion methods, DeCo is more efficient and effective than the multi-scale cascade model PixelFlow~\cite{pixelflow}, which suffers from high computational costs. DeCo also shows superior performance compared to the single-scale attention-based PixDDT~\cite{decoupled_dit}. Compared to the recent PixNerd~\cite{wang2025pixnerd}, our method achieves a better FID with lower training and inference costs.

JiT identifies that the high-dimensional noise may distract the model with limited capacity from learning low-dimensional data~\cite{li2025jit}. To address this, it predicts the clean image and anchors the generation process to the low-dimensional data manifold, successfully reducing the FID from 61.10 to 39.06, as shown in \cref{tab:baseline_comparison}.
\emph{Our DeCo shares the similar motivation, i.e., preventing high-frequency signals with high-dimensional noise from interfering with the DiT's ability to learn low-frequency semantics.} However, our DeCo proposes an alternative architectural solution. We introduce a lightweight pixel decoder to focus on modeling high-frequency signals and free the DiT to learn low-frequency semantics. Our DeCo can also alleviate the negative impact of the high-frequency noise in the clean image, such as camera noise.
Consequently, our DeCo achieves a superior FID of 31.35 compared to the 39.06 of JiT.

\subsection{Class-to-Image Generation}
\noindent\textbf{Setup.}
Please refer to Appendix~\ref{appendix: c2i_train_details} for detailed setups.

\noindent\textbf{Main Results.}
Our DeCo achieves leading  FID of 1.62 on ImageNet 256$\times$256 and 2.22 on ImageNet 512$\times$512.
At the 256$\times$256 resolution, DeCo demonstrates remarkable inference efficiency. It generates an image in just 1.05s with 100 inference steps, whereas RDM~\cite{rdm} requires 38.4s and PixelFlow~\cite{pixelflow} needs 9.78s. In terms of training efficiency, as shown in \cref{tab:baseline_comparison}, a single iteration for our model takes only 0.24s, far less than PixelFlow's 1.61s. 
When trained for the same 320 epochs, our model's FID (1.90) is substantially lower than the baseline's 2.79 and surpasses the recent PixelFlow and PixNerd.
As illustrated in \cref{fig:fid_comparison}, DeCo achieves a FID of 2.57 in just 80 epochs (400k iterations), which exceeds the baseline's FID at 800 epochs, marking a 10$\times$ improvement in training efficiency. After 800 training epochs, our DeCo achieves a superior FID of 1.62 with 250 sampling steps across pixel diffusion models, which is even comparable to the two-stage latent diffusion models. Using the same heun~\cite{heun1900neue} sampler and 50-step inference at 600 epochs, DeCo reaches an FID of 1.69, outperforming JiT's FID of 1.86 with fewer parameters and FLOPs.
At the 512$\times$512 resolution, our DeCo model substantially outperforms existing pixel-based diffusion methods, setting a leading FID of 2.22. Moreover, by fine-tuning our ImageNet 256$\times$256 model at 320 epochs for 20 additional epochs following PixNerd~\cite{wang2025pixnerd}, our FID and IS are comparable to those of DiT-XL/2~\cite{dit} and SiT-XL/2~\cite{sit} after 600 training epochs.

\subsection{Text-to-Image Generation}

\noindent\textbf{Setup.}
Please refer to Appendix~\ref{appendix: t2i_train_details} for training details.


\noindent\textbf{Main Results.}
Compared to two-stage latent diffusion methods, our DeCo achieves an overall score of 0.86 on the GenEval benchmark~\cite{geneval}. This result outperforms prominent text-to-image models such as SD3~\cite{sd3} and FLUX.1-dev~\cite{flux2024}, as well as unified models including BLIP3o~\cite{blip3o} and OmniGen2~\cite{wu2025omnigen2}. Notably, our model achieves superior performance despite using the same training data as BLIP3o. On DPG-Bench~\cite{dpg}, DeCo delivers a competitive average score comparable to two-stage latent diffusion methods.
When compared to other end-to-end pixel diffusion methods, DeCo achieves a significant performance advantage over PixelFlow and PixNerd. 
These results show that end-to-end pixel diffusion, as implemented in our DeCo, can achieve performance comparable to that of two-stage methods with limited training and inference costs.
Visualizations of images generated by our text-to-image DeCo can be found in \cref{fig:t2i_visualization}. Please see Appendix~\ref{appendix:textual_prompts} for prompts.

\subsection{More Ablations}
This section presents ablation studies on the pixel decoder design, the interaction mechanism between the DiT and the pixel decoder, and the hyperparameters of the frequency-aware FM loss. All experiments follow the setup in \cref{sec:exp_baseline}.

\noindent\textbf{Hidden Size of Pixel Decoder.}
As shown in \cref{exp: tab_ablation}~(a), DeCo achieves the best performance when the hidden size $d$ is set to 32. Smaller sizes limit model capacity, while larger sizes offer no further gains. Thus, we use a hidden size of 32 by default.

\noindent\textbf{Depth of Pixel Decoder.}
In \cref{exp: tab_ablation}~(b), a 3-layer decoder achieves the best results. A single layer lacks capacity, whereas a 6-layer design may introduce optimization difficulties. With a hidden size of 32 and 3 layers, our attention-free decoder is lightweight (8.5M parameters) and efficient for high-resolution inputs.

\noindent\textbf{Patch Size of Pixel Decoder.}
As shown in \cref{exp: tab_ablation}~(c), DeCo performs best when the decoder's patch size is set to 1, enabling direct processing of the full-resolution input. Patchifying the decoder's input degrades results, with the worst performance observed when using a large patch size of 16 like DiT. This demonstrates the effectiveness of our multi-scale input strategy.
All comparisons use similar parameter counts and computational costs.


\noindent\textbf{Interaction between DiT and Pixel Decoder.}
\cref{exp: tab_ablation}~(d) shows that simply upsampling DiT outputs and adding them to dense decoder features, as done in UNet~\cite{ronneberger2015u}, underperforms compared to AdaLN-based interaction.
AdaLN~\cite{dit} provides a more effective interaction mechanism, using the DiT output as a semantic condition for velocity prediction.

\noindent\textbf{Loss Weight.}
In \cref{exp: tab_ablation}~(e), the loss weight of 1 for $\mathcal{L}_\mathrm{FreqFM}$ gives the best results, which we adopt as the default setting.

\noindent\textbf{JPEG Quality in $\mathcal{L}_\mathrm{FreqFM}$.}
In \cref{exp: tab_ablation}~(f), we study the effect of the JPEG quality factor in $\mathcal{L}_\mathrm{FreqFM}$. With a quality of 100 (lossless compression), all frequency components are equally weighted, yielding an FID of 33.84, close to 34.12 without $\mathcal{L}_\mathrm{FreqFM}$ as expected. The commonly used quality of 85 performs best, emphasizing the important frequencies while slightly downweighting insignificant ones for optimal balance. Reducing quality to 50 overly suppresses high-frequency signals, slightly harming performance. Therefore, we use a JPEG quality of 85 in all experiments.

\section{Conclusions}
We introduced DeCo, a novel frequency-decoupled framework for pixel diffusion. By separating the modeling of low-frequency semantics with a DiT and high-frequency signals with a lightweight pixel decoder, DeCo significantly improves generation quality and efficiency. Our proposed frequency-aware FM loss further enhances visual quality by prioritizing perceptually important frequencies. DeCo achieves leading performance in pixel diffusion on both class-to-image and text-to-image generation benchmarks, closing the gap with two-stage latent diffusion methods.


\section*{Acknowledgments} 
This work was supported in part by the Grant 2023-JCJQ-LA-001-088, in part by the Grant 2025ZD1601300, in part by the Natural Science Foundation of China under Grant U20B2052, in part by the Okawa Foundation Research Award, in part by the Ant Group Research Fund, and in part by the Kunpeng\&Ascend Center of Excellence, Peking University.

{
    \small
    \bibliographystyle{ieeenat_fullname}
    \bibliography{main}

@String(AAAI = {AAAI})

@misc{chen2023pixartalpha,
      title={PixArt-$\alpha$: Fast Training of Diffusion Transformer for Photorealistic Text-to-Image Synthesis}, 
      author={Junsong Chen and Jincheng Yu and Chongjian Ge and Lewei Yao and Enze Xie and Yue Wu and Zhongdao Wang and James Kwok and Ping Luo and Huchuan Lu and Zhenguo Li},
      year={2023},
      eprint={2310.00426},
      archivePrefix={arXiv},
      primaryClass={cs.CV}
}

@inproceedings{
lipman2023flow,
title={Flow Matching for Generative Modeling},
author={Yaron Lipman and Ricky T. Q. Chen and Heli Ben-Hamu and Maximilian Nickel and Matthew Le},
booktitle={The Eleventh International Conference on Learning Representations },
year={2023},
url={https://openreview.net/forum?id=PqvMRDCJT9t}
}

@misc{
luo2024latent,
title={Latent Consistency Models: Synthesizing High-Resolution Images with Few-step Inference},
author={Simian Luo and Yiqin Tan and Longbo Huang and Jian Li and Hang Zhao},
year={2024},
url={https://openreview.net/forum?id=duBCwjb68o}
}

@article{wu2024videomaker,
  title={Videomaker: Zero-shot customized video generation with the inherent force of video diffusion models},
  author={Wu, Tao and Zhang, Yong and Cun, Xiaodong and Qi, Zhongang and Pu, Junfu and Dou, Huanzhang and Zheng, Guangcong and Shan, Ying and Li, Xi},
  journal={arXiv preprint arXiv:2412.19645},
  year={2024}
}

@inproceedings{wu2025customcrafter,
  title={Customcrafter: Customized video generation with preserving motion and concept composition abilities},
  author={Wu, Tao and Zhang, Yong and Wang, Xintao and Zhou, Xianpan and Zheng, Guangcong and Qi, Zhongang and Shan, Ying and Li, Xi},
  booktitle={Proceedings of the AAAI Conference on Artificial Intelligence},
  volume={39},
  number={8},
  pages={8469--8477},
  year={2025}
}

@inproceedings{wu2024spherediffusion,
  title={Spherediffusion: Spherical geometry-aware distortion resilient diffusion model},
  author={Wu, Tao and Li, Xuewei and Qi, Zhongang and Hu, Di and Wang, Xintao and Shan, Ying and Li, Xi},
  booktitle={Proceedings of the AAAI Conference on Artificial Intelligence},
  volume={38},
  number={6},
  pages={6126--6134},
  year={2024}
}

@misc{flux2024,
    author={Black Forest Labs},
    title={FLUX},
    year={2024},
    howpublished={\url{https://github.com/black-forest-labs/flux}},
}

@article{wu2025omnigen2,
  title={OmniGen2: Exploration to Advanced Multimodal Generation},
  author={Chenyuan Wu and Pengfei Zheng and Ruiran Yan and Shitao Xiao and Xin Luo and Yueze Wang and Wanli Li and Xiyan Jiang and Yexin Liu and Junjie Zhou and Ze Liu and Ziyi Xia and Chaofan Li and Haoge Deng and Jiahao Wang and Kun Luo and Bo Zhang and Defu Lian and Xinlong Wang and Zhongyuan Wang and Tiejun Huang and Zheng Liu},
  journal={arXiv preprint arXiv:2506.18871},
  year={2025}
}

@article{blip3o,
  title={Blip3-o: A family of fully open unified multimodal models-architecture, training and dataset},
  author={Chen, Jiuhai and Xu, Zhiyang and Pan, Xichen and Hu, Yushi and Qin, Can and Goldstein, Tom and Huang, Lifu and Zhou, Tianyi and Xie, Saining and Savarese, Silvio and others},
  journal={arXiv preprint arXiv:2505.09568},
  year={2025}
}

@article{geneval,
  title={Geneval: An object-focused framework for evaluating text-to-image alignment},
  author={Ghosh, Dhruba and Hajishirzi, Hannaneh and Schmidt, Ludwig},
  journal={Advances in Neural Information Processing Systems},
  volume={36},
  pages={52132--52152},
  year={2023}
}

@article{dpg,
  title={Ella: Equip diffusion models with llm for enhanced semantic alignment},
  author={Hu, Xiwei and Wang, Rui and Fang, Yixiao and Fu, Bin and Cheng, Pei and Yu, Gang},
  journal={arXiv preprint arXiv:2403.05135},
  year={2024}
}

@article{dmm,
  title={Dmm: Building a versatile image generation model via distillation-based model merging},
  author={Song, Tianhui and Feng, Weixin and Wang, Shuai and Li, Xubin and Ge, Tiezheng and Zheng, Bo and Wang, Limin},
  journal={arXiv preprint arXiv:2504.12364},
  year={2025}
}

@article{seedream2,
  title={Seedream 2.0: A native chinese-english bilingual image generation foundation model},
  author={Gong, Lixue and Hou, Xiaoxia and Li, Fanshi and Li, Liang and Lian, Xiaochen and Liu, Fei and Liu, Liyang and Liu, Wei and Lu, Wei and Shi, Yichun and others},
  journal={arXiv preprint arXiv:2503.07703},
  year={2025}
}

@article{seedream3,
  title={Seedream 3.0 technical report},
  author={Gao, Yu and Gong, Lixue and Guo, Qiushan and Hou, Xiaoxia and Lai, Zhichao and Li, Fanshi and Li, Liang and Lian, Xiaochen and Liao, Chao and Liu, Liyang and others},
  journal={arXiv preprint arXiv:2504.11346},
  year={2025}
}

@article{mogao,
  title={Mogao: An Omni Foundation Model for Interleaved Multi-Modal Generation},
  author={Liao, Chao and Liu, Liyang and Wang, Xun and Luo, Zhengxiong and Zhang, Xinyu and Zhao, Wenliang and Wu, Jie and Li, Liang and Tian, Zhi and Huang, Weilin},
  journal={arXiv preprint arXiv:2505.05472},
  year={2025}
}

@inproceedings{sid,
  title={Score identity distillation: Exponentially fast distillation of pretrained diffusion models for one-step generation},
  author={Zhou, Mingyuan and Zheng, Huangjie and Wang, Zhendong and Yin, Mingzhang and Huang, Hai},
  booktitle={Forty-first International Conference on Machine Learning},
  year={2024}
}

@inproceedings{simple_diffusion,
  title={simple diffusion: End-to-end diffusion for high resolution images},
  author={Hoogeboom, Emiel and Heek, Jonathan and Salimans, Tim},
  booktitle={International Conference on Machine Learning},
  pages={13213--13232},
  year={2023},
  organization={PMLR}
}

@article{vdm,
  title={Understanding diffusion objectives as the elbo with simple data augmentation},
  author={Kingma, Diederik and Gao, Ruiqi},
  journal={Advances in Neural Information Processing Systems},
  volume={36},
  pages={65484--65516},
  year={2023}
}

@article{jetformer,
  title={JetFormer: An autoregressive generative model of raw images and text},
  author={Tschannen, Michael and Pinto, Andr{\'e} Susano and Kolesnikov, Alexander},
  journal={arXiv preprint arXiv:2411.19722},
  year={2024}
}

@article{pr_recall,
  title={Improved precision and recall metric for assessing generative models},
  author={Kynk{\"a}{\"a}nniemi, Tuomas and Karras, Tero and Laine, Samuli and Lehtinen, Jaakko and Aila, Timo},
  journal={Advances in neural information processing systems},
  volume={32},
  year={2019}
}

@article{is,
  title={Improved techniques for training gans},
  author={Salimans, Tim and Goodfellow, Ian and Zaremba, Wojciech and Cheung, Vicki and Radford, Alec and Chen, Xi},
  journal={Advances in neural information processing systems},
  volume={29},
  year={2016}
}

@article{sfid,
  title={Generating images with sparse representations},
  author={Nash, Charlie and Menick, Jacob and Dieleman, Sander and Battaglia, Peter W},
  journal={arXiv preprint arXiv:2103.03841},
  year={2021}
}

@article{fid,
  title={Gans trained by a two time-scale update rule converge to a local nash equilibrium},
  author={Heusel, Martin and Ramsauer, Hubert and Unterthiner, Thomas and Nessler, Bernhard and Hochreiter, Sepp},
  journal={Advances in neural information processing systems},
  volume={30},
  year={2017}
}

@article{cfg,
  title={Classifier-free diffusion guidance},
  author={Ho, Jonathan and Salimans, Tim},
  journal={arXiv preprint arXiv:2207.12598},
  year={2022}
}

@article{interval_guidance,
  title={Applying guidance in a limited interval improves sample and distribution quality in diffusion models},
  author={Kynk{\"a}{\"a}nniemi, Tuomas and Aittala, Miika and Karras, Tero and Laine, Samuli and Aila, Timo and Lehtinen, Jaakko},
  journal={arXiv preprint arXiv:2404.07724},
  year={2024}
}

@article{ddpm,
  title={Denoising diffusion probabilistic models},
  author={Ho, Jonathan and Jain, Ajay and Abbeel, Pieter},
  journal={Advances in neural information processing systems},
  volume={33},
  pages={6840--6851},
  year={2020}
}

@article{ddim,
  title={Denoising Diffusion Implicit Models},
  author={Song, Jiaming and Meng, Chenlin and Ermon, Stefano},
  journal={arXiv:2010.02502},
  year={2020},
  month={October},
  abbr={Preprint},
  url={https://arxiv.org/abs/2010.02502}
}

@article{adm,
  title={Diffusion models beat gans on image synthesis},
  author={Dhariwal, Prafulla and Nichol, Alexander},
  journal={Advances in neural information processing systems},
  volume={34},
  pages={8780--8794},
  year={2021}
}

@article{rin,
  title={Scalable adaptive computation for iterative generation},
  author={Jabri, Allan and Fleet, David and Chen, Ting},
  journal={arXiv preprint arXiv:2212.11972},
  year={2022}
}

@inproceedings{uvit,
  title={All are worth words: A vit backbone for diffusion models},
  author={Bao, Fan and Nie, Shen and Xue, Kaiwen and Cao, Yue and Li, Chongxuan and Su, Hang and Zhu, Jun},
  booktitle={Proceedings of the IEEE/CVF Conference on Computer Vision and Pattern Recognition},
  pages={22669--22679},
  year={2023}
}

@inproceedings{dit,
  title={Scalable diffusion models with transformers},
  author={Peebles, William and Xie, Saining},
  booktitle={Proceedings of the IEEE/CVF International Conference on Computer Vision},
  pages={4195--4205},
  year={2023}
}

@inproceedings{edm2,
  title={Analyzing and improving the training dynamics of diffusion models},
  author={Karras, Tero and Aittala, Miika and Lehtinen, Jaakko and Hellsten, Janne and Aila, Timo and Laine, Samuli},
  booktitle={Proceedings of the IEEE/CVF Conference on Computer Vision and Pattern Recognition},
  pages={24174--24184},
  year={2024}
}

@article{decoupled_dit,
  title={Decoupled diffusion transformer},
  author={Wang, Shuai and Tian, Zhi and Huang, Weilin and Wang, Limin},
  journal={arXiv preprint arXiv:2504.05741},
  year={2025}
}

@article{repa_e,
  title={REPA-E: Unlocking VAE for End-to-End Tuning with Latent Diffusion Transformers},
  author={Leng, Xingjian and Singh, Jaskirat and Hou, Yunzhong and Xing, Zhenchang and Xie, Saining and Zheng, Liang},
  journal={arXiv preprint arXiv:2504.10483},
  year={2025}
}

@article{pixelflow,
  title={PixelFlow: Pixel-Space Generative Models with Flow},
  author={Chen, Shoufa and Ge, Chongjian and Zhang, Shilong and Sun, Peize and Luo, Ping},
  journal={arXiv preprint arXiv:2504.07963},
  year={2025}
}

@article{dod,
  title={Diffusion models need visual priors for image generation},
  author={Yue, Xiaoyu and Wang, Zidong and Lu, Zeyu and Sun, Shuyang and Wei, Meng and Ouyang, Wanli and Bai, Lei and Zhou, Luping},
  journal={arXiv preprint arXiv:2410.08531},
  year={2024}
}

@article{dim,
  title={Dim: Diffusion mamba for efficient high-resolution image synthesis},
  author={Teng, Yao and Wu, Yue and Shi, Han and Ning, Xuefei and Dai, Guohao and Wang, Yu and Li, Zhenguo and Liu, Xihui},
  journal={arXiv preprint arXiv:2405.14224},
  year={2024}
}

@article{vavae,
  title={Reconstruction vs. Generation: Taming Optimization Dilemma in Latent Diffusion Models},
  author={Yao, Jingfeng and Wang, Xinggang},
  journal={arXiv preprint arXiv:2501.01423},
  year={2025}
}

@article{dcae,
  title={Deep compression autoencoder for efficient high-resolution diffusion models},
  author={Chen, Junyu and Cai, Han and Chen, Junsong and Xie, Enze and Yang, Shang and Tang, Haotian and Li, Muyang and Lu, Yao and Han, Song},
  journal={arXiv preprint arXiv:2410.10733},
  year={2024}
}

@article{sit,
  title={SiT: Exploring Flow and Diffusion-based Generative Models with Scalable Interpolant Transformers},
  author={Ma, Nanye and Goldstein, Mark and Albergo, Michael S and Boffi, Nicholas M and Vanden-Eijnden, Eric and Xie, Saining},
  journal={arXiv preprint arXiv:2401.08740},
  year={2024}
}

@article{fractal,
  title={Fractal generative models},
  author={Li, Tianhong and Sun, Qinyi and Fan, Lijie and He, Kaiming},
  journal={arXiv preprint arXiv:2502.17437},
  year={2025}
}

@article{tarflow,
  title={Normalizing flows are capable generative models},
  author={Zhai, Shuangfei and Zhang, Ruixiang and Nakkiran, Preetum and Berthelot, David and Gu, Jiatao and Zheng, Huangjie and Chen, Tianrong and Bautista, Miguel Angel and Jaitly, Navdeep and Susskind, Josh},
  journal={arXiv preprint arXiv:2412.06329},
  year={2024}
}

@article{fluid,
  title={Fluid: Scaling autoregressive text-to-image generative models with continuous tokens},
  author={Fan, Lijie and Li, Tianhong and Qin, Siyang and Li, Yuanzhen and Sun, Chen and Rubinstein, Michael and Sun, Deqing and He, Kaiming and Tian, Yonglong},
  journal={arXiv preprint arXiv:2410.13863},
  year={2024}
}

@article{rdm,
  title={Relay diffusion: Unifying diffusion process across resolutions for image synthesis},
  author={Teng, Jiayan and Zheng, Wendi and Ding, Ming and Hong, Wenyi and Wangni, Jianqiao and Yang, Zhuoyi and Tang, Jie},
  journal={arXiv preprint arXiv:2309.03350},
  year={2023}
}

@article{repa,
  title={Representation alignment for generation: Training diffusion transformers is easier than you think},
  author={Yu, Sihyun and Kwak, Sangkyung and Jang, Huiwon and Jeong, Jongheon and Huang, Jonathan and Shin, Jinwoo and Xie, Saining},
  journal={arXiv preprint arXiv:2410.06940},
  year={2024}
}

@article{dinov2,
  title={Dinov2: Learning robust visual features without supervision},
  author={Oquab, Maxime and Darcet, Timoth{\'e}e and Moutakanni, Th{\'e}o and Vo, Huy and Szafraniec, Marc and Khalidov, Vasil and Fernandez, Pierre and Haziza, Daniel and Massa, Francisco and El-Nouby, Alaaeldin and others},
  journal={arXiv preprint arXiv:2304.07193},
  year={2023}
}

@article{sd3,
  title={Scaling rectified flow transformers for high-resolution image synthesis},
  author={Esser, Patrick and Kulal, Sumith and Blattmann, Andreas and Entezari, Rahim and M{\"u}ller, Jonas and Saini, Harry and Levi, Yam and Lorenz, Dominik and Sauer, Axel and Boesel, Frederic and others},
  journal={arXiv preprint arXiv:2403.03206},
  year={2024}
}

@inproceedings{ldm,
  title={High-resolution image synthesis with latent diffusion models},
  author={Rombach, Robin and Blattmann, Andreas and Lorenz, Dominik and Esser, Patrick and Ommer, Bj{\"o}rn},
  booktitle={Proceedings of the IEEE/CVF conference on computer vision and pattern recognition},
  pages={10684--10695},
  year={2022}
}

@article{rope,
  title={Roformer: Enhanced transformer with rotary position embedding},
  author={Su, Jianlin and Ahmed, Murtadha and Lu, Yu and Pan, Shengfeng and Bo, Wen and Liu, Yunfeng},
  journal={Neurocomputing},
  volume={568},
  pages={127063},
  year={2024},
  publisher={Elsevier}
}

@article{flowdcn,
  title={Exploring DCN-like architecture for fast image generation with arbitrary resolution},
  author={Wang, Shuai and Li, Zexian and Song, Tianhui and Li, Xubin and Ge, Tiezheng and Zheng, Bo and Wang, Limin},
  journal={Advances in Neural Information Processing Systems},
  volume={37},
  pages={87959--87977},
  year={2024}
}

@article{nit,
  title={Native-Resolution Image Synthesis},
  author={Wang, Zidong and Bai, Lei and Yue, Xiangyu and Ouyang, Wanli and Zhang, Yiyuan},
  journal={arXiv preprint arXiv:2506.03131},
  year={2025}
}

@article{llama1,
  title={Llama: Open and efficient foundation language models},
  author={Touvron, Hugo and Lavril, Thibaut and Izacard, Gautier and Martinet, Xavier and Lachaux, Marie-Anne and Lacroix, Timoth{\'e}e and Rozi{\`e}re, Baptiste and Goyal, Naman and Hambro, Eric and Azhar, Faisal and others},
  journal={arXiv preprint arXiv:2302.13971},
  year={2023}
}

@article{llama2,
  title={Llama 2: Open foundation and fine-tuned chat models},
  author={Touvron, Hugo and Martin, Louis and Stone, Kevin and Albert, Peter and Almahairi, Amjad and Babaei, Yasmine and Bashlykov, Nikolay and Batra, Soumya and Bhargava, Prajjwal and Bhosale, Shruti and others},
  journal={arXiv preprint arXiv:2307.09288},
  year={2023}
}

@article{wang2025pixnerd,
  title={Pixnerd: Pixel neural field diffusion},
  author={Wang, Shuai and Gao, Ziteng and Zhu, Chenhui and Huang, Weilin and Wang, Limin},
  journal={arXiv preprint arXiv:2507.23268},
  year={2025}
}

@article{yang2025qwen3,
  title={Qwen3 technical report},
  author={Yang, An and Li, Anfeng and Yang, Baosong and Zhang, Beichen and Hui, Binyuan and Zheng, Bo and Yu, Bowen and Gao, Chang and Huang, Chengen and Lv, Chenxu and others},
  journal={arXiv preprint arXiv:2505.09388},
  year={2025}
}

@misc{Betker2023ImprovingIG,
  title        = {Improving Image Generation with Better Captions},
  author       = {Betker, James and Goh, Gabriel and Jing, Li and Brooks, Tim and Wang, Jianfeng and Li, Linjie and Ouyang, Long and Zhuang, Juntang and Lee, Joyce and Guo, Yufei and Manassra, Wesam and Dhariwal, Prafulla and Chu, Casey and Jiao, Yunxin and Ramesh, Aditya},
  howpublished = {OpenAI Technical Report},
  year         = {2023},
  url          = {https://cdn.openai.com/papers/dall-e-3.pdf}
}

@techreport{JPEG,
  title = {Information technology — Digital compression and coding of continuous-tone still images: Requirements and guidelines},
  author = "{Joint Photographic Experts Group}",
  number = {ITU-T T.81},
  institution = {International Telecommunication Union (ITU-T)},
  year = {1992},
  url = {https://www.itu.int/rec/T-REC-T.81-199209-I/en}
}

@article{elfwing2018sigmoid,
  title={Sigmoid-weighted linear units for neural network function approximation in reinforcement learning},
  author={Elfwing, Stefan and Uchibe, Eiji and Doya, Kenji},
  journal={Neural networks},
  volume={107},
  pages={3--11},
  year={2018},
  publisher={Elsevier}
}

@article{si2022inception,
  title={Inception transformer},
  author={Si, Chenyang and Yu, Weihao and Zhou, Pan and Zhou, Yichen and Wang, Xinchao and Yan, Shuicheng},
  journal={Advances in Neural Information Processing Systems},
  volume={35},
  pages={23495--23509},
  year={2022}
}

@inproceedings{chenvision,
  title={Vision Transformer Adapter for Dense Predictions},
  author={Chen, Zhe and Duan, Yuchen and Wang, Wenhai and He, Junjun and Lu, Tong and Dai, Jifeng and Qiao, Yu},
  booktitle={The Eleventh International Conference on Learning Representations},
  year={2023}
}

@inproceedings{
park2022how,
title={How Do Vision Transformers Work?},
author={Namuk Park and Songkuk Kim},
booktitle={International Conference on Learning Representations},
year={2022},
url={https://openreview.net/forum?id=D78Go4hVcxO}
}

@article{highfreq_gan,
author = {Li, Ziqiang and Xia, Pengfei and Rui, Xue and Li, Bin},
title = {Exploring the Effect of High-frequency Components in GANs Training},
year = {2023},
issue_date = {September 2023},
publisher = {Association for Computing Machinery},
address = {New York, NY, USA},
volume = {19},
number = {5},
issn = {1551-6857},
url = {https://doi.org/10.1145/3578585},
doi = {10.1145/3578585},
journal = {ACM Trans. Multimedia Comput. Commun. Appl.},
month = mar,
articleno = {153},
numpages = {22}
}

@article{wang2022fregan,
  title={FreGAN: Exploiting frequency components for training GANs under limited data},
  author={Wang, Zhe and Chi, Ziqiu and Zhang, Yanbing and others},
  journal={Advances in Neural Information Processing Systems},
  volume={35},
  pages={33387--33399},
  year={2022}
}

@inproceedings{
skorokhodov2025improving,
title={Improving the Diffusability of Autoencoders},
author={Ivan Skorokhodov and Sharath Girish and Benran Hu and Willi Menapace and Yanyu Li and Rameen Abdal and Sergey Tulyakov and Aliaksandr Siarohin},
booktitle={Forty-second International Conference on Machine Learning},
year={2025},
url={https://openreview.net/forum?id=2hEDcA7xy4}
}

@article{denton2015deep_multi_reso,
  title={Deep generative image models using a laplacian pyramid of adversarial networks},
  author={Denton, Emily L and Chintala, Soumith and Fergus, Rob and others},
  journal={Advances in neural information processing systems},
  volume={28},
  year={2015}
}

@inproceedings{karras2018progressive_multi_reso,
  title={Progressive Growing of GANs for Improved Quality, Stability, and Variation},
  author={Karras, Tero and Aila, Timo and Laine, Samuli and Lehtinen, Jaakko},
  booktitle={International Conference on Learning Representations},
  year={2018}
}

@inproceedings{wang2020high,
  title={High-frequency component helps explain the generalization of convolutional neural networks},
  author={Wang, Haohan and Wu, Xindi and Huang, Zeyi and Xing, Eric P},
  booktitle={Proceedings of the IEEE/CVF conference on computer vision and pattern recognition},
  pages={8684--8694},
  year={2020}
}

@inproceedings{ronneberger2015u,
  title={U-net: Convolutional networks for biomedical image segmentation},
  author={Ronneberger, Olaf and Fischer, Philipp and Brox, Thomas},
  booktitle={International Conference on Medical image computing and computer-assisted intervention},
  pages={234--241},
  year={2015},
  organization={Springer}
}

@inproceedings{
loshchilov2018decoupled,
title={Decoupled Weight Decay Regularization},
author={Ilya Loshchilov and Frank Hutter},
booktitle={International Conference on Learning Representations},
year={2019},
url={https://openreview.net/forum?id=Bkg6RiCqY7},
}

@book{pennebaker1992jpeg,
  title={JPEG: Still image data compression standard},
  author={Pennebaker, William B and Mitchell, Joan L},
  year={1992},
  publisher={Springer Science \& Business Media}
}

@misc{li2025jit,
      title={Back to Basics: Let Denoising Generative Models Denoise}, 
      author={Tianhong Li and Kaiming He},
      year={2025},
      eprint={2511.13720},
      archivePrefix={arXiv},
      primaryClass={cs.CV},
      url={https://arxiv.org/abs/2511.13720}, 
}

@article{heun1900neue,
  title={Neue Methoden zur approximativen Integration der Differentialgleichungen einer unabh{\"a}ngigen Ver{\"a}nderlichen},
  author={Heun, Karl and others},
  journal={Z. Math. Phys},
  volume={45},
  pages={23--38},
  year={1900}
}

@article{ma2026pixelgen,
  title={PixelGen: Pixel Diffusion Beats Latent Diffusion with Perceptual Loss},
  author={Ma, Zehong and Xu, Ruihan and Zhang, Shiliang},
  journal={arXiv preprint arXiv:2602.02493},
  year={2026}
}

@inproceedings{
ma2025magcache,
title={MagCache: Fast Video Generation with Magnitude-Aware Cache},
author={Zehong Ma and Longhui Wei and Feng Wang and Shiliang Zhang and Qi Tian},
booktitle={The Thirty-ninth Annual Conference on Neural Information Processing Systems},
year={2025},
url={https://openreview.net/forum?id=KZn7TDOL4J}
}

@article{wu2025multicrafter,
  title={MultiCrafter: High-Fidelity Multi-Subject Generation via Disentangled Attention and Identity-Aware Preference Alignment},
  author={Wu, Tao and Jiang, Yibo and Lu, Yehao and Wang, Zhizhong and Huang, Zeyi and Qin, Zequn and Li, Xi},
  journal={arXiv preprint arXiv:2509.21953},
  year={2025}
}

@misc{tian2024udits,
      title={U-DiTs: Downsample Tokens in U-Shaped Diffusion Transformers}, 
      author={Yuchuan Tian and Zhijun Tu and Hanting Chen and Jie Hu and Chao Xu and Yunhe Wang},
      year={2024},
      eprint={2405.02730},
      archivePrefix={arXiv},
      primaryClass={cs.CV}
}

@article{tian2025dic,
  author       = {Yuchuan Tian and
                  Jing Han and
                  Chengcheng Wang and
                  Yuchen Liang and
                  Chao Xu and
                  Hanting Chen},
  title        = {DiC: Rethinking Conv3x3 Designs in Diffusion Models},
  journal      = {CoRR},
  volume       = {abs/2501.00603},
  year         = {2025},
  url          = {https://doi.org/10.48550/arXiv.2501.00603},
  doi          = {10.48550/ARXIV.2501.00603},
  eprinttype    = {arXiv},
  eprint       = {2501.00603},
  bibsource    = {dblp computer science bibliography, https://dblp.org}
}

@inproceedings{
tian2025urepa,
title={U-REPA: Aligning Diffusion U-Nets to ViTs},
author={Yuchuan Tian and Hanting Chen and Mengyu Zheng and Yuchen Liang and Chao Xu and Yunhe Wang},
booktitle={The Thirty-ninth Annual Conference on Neural Information Processing Systems},
year={2025},
url={https://openreview.net/forum?id=im3FJ6quii}
}

@article{wang2025mosa,
  title={MoSA: Motion-Coherent Human Video Generation via Structure-Appearance Decoupling},
  author={Wang, Haoyu and Tang, Hao and Di, Donglin and Zhang, Zhilu and Zuo, Wangmeng and Gao, Feng and Ma, Siwei and Zhang, Shiliang},
  journal={arXiv preprint arXiv:2508.17404},
  year={2025}
}

@inproceedings{wang2025mv,
  title={Mv-vton: Multi-view virtual try-on with diffusion models},
  author={Wang, Haoyu and Zhang, Zhilu and Di, Donglin and Zhang, Shiliang and Zuo, Wangmeng},
  booktitle={Proceedings of the AAAI Conference on Artificial Intelligence},
  volume={39},
  number={7},
  pages={7682--7690},
  year={2025}
}
}
\clearpage
\setcounter{page}{1}
\maketitlesupplementary
\appendix

\section{Comparison with JiT}
\label{appendix: comparison_with_jit}
\begin{figure}[h!]
    \vspace{-9pt}
    \centering
    \includegraphics[width=1.0\linewidth]{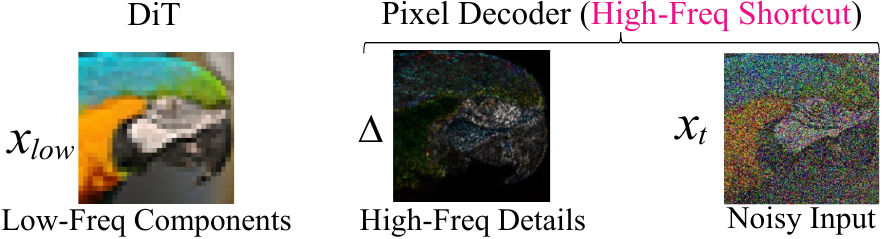}
    \caption{Comparison with JiT.}
    \vspace{-9pt}
    \label{fig: comparison_with_jit}
\end{figure}

\noindent The key idea of DeCo is to provide \textcolor{RubineRed}{{a high-freq shortcut via the pixel decoder}}, enabling an implicit frequency decoupling. High-freq components include noisy input $x_t$ and high-freq details $\Delta$. In standard DiTs, high-freq signals tend to be suppressed after the patchification and deep transformer layers. Since pixel decoder directly models each pixel of the raw input, DeCo can automatically use this high-freq shortcut and offload high-freq modeling to pixel decoder, enabling DiT to focus on low-freq components $x_{low}$. 
Unlike JiT, which explicitly predicts $x_0=x_{low}+\Delta$, our DiT implicitly models $x_{low}$, excluding hard high-freq details $\Delta$. The pixel decoder, guided by $x_{low}$, only needs to fit a simple JiT-style formulation $v_\theta = \frac{x_{low}+(\Delta-x_t)}{t}$. This separation is learned end-to-end. Visualization in ~\cref{fig:intro}(c), spectrum analysis in \cref{fig:frequency_plot}, and superior performance over ``JiT+REPA'' in \cref{tab:baseline_comparison} validate our implicit decoupling. 

\section{Implementary Details}
\subsection{Baseline Comparisons}
\label{appendix: baseline_train_details}
In this subsection, we summarize the settings used for all baseline comparisons.
In the baseline comparisons, all diffusion models are trained on ImageNet at 256$\times$256 resolution for 200k iterations using a large DiT variant. Following previous works~\cite{dit,wang2025pixnerd}, we use a global batch size of 256 and the AdamW optimizer with a constant learning rate of 1e-4. 
Both baseline and DeCo adopt SwiGLU~\cite{llama2, llama1}, RoPE2d~\cite{rope}, and RMSNorm, and are trained with lognorm sampling and REPA~\cite{repa}.
The patch size of DiT's input is set to 16 for both baseline and our DeCo.
The patch size of pixel decoder is set to 1.
Our main architectural change on the baseline is to replace the final two DiT blocks of the baseline with our proposed pixel decoder. 

For inference, we use 50 Euler steps without classifier-free guidance~\cite{cfg} (CFG) for all models except PixelFlow~\cite{pixelflow}, which requires 100 steps.
We also report results for the two-stage DiT-L/2 that requires a VAE and the recent pixel diffusion models PixelFlow~\cite{pixelflow} and PixNerd~\cite{wang2025pixnerd}. For a fair comparison, we further integrate DDT~\cite{decoupled_dit} into the pixel diffusion to form PixDDT, which has the similar parameter counts and computation FLOPs to our DeCo.
Training memory and speed are measured with batch size of 256 on 8$\times$A800 GPUs, while inference memory and time are measured on a single A800 with batch size of 1.

\subsection{Class-to-Image Generation}
\label{appendix: c2i_train_details}
For class-to-image generation experiments on ImageNet, we first train the model at a 256$\times$256 resolution for 320 epochs. Subsequently, we fine-tune the model for additional 20 epochs at a 512$\times$512 resolution. During inference, we use 100 Euler steps with CFG~\cite{cfg} and guidance interval~\cite{interval_guidance}. Inference latency is measured on a single A800 GPU.
The batch size and learning rate follow the default settings previously described. We use a global batch size of 256 and the AdamW optimizer with a constant learning rate of 1e-4.  We set the CFG scale to 3.2 for the $256\times256$ resolution (320 epochs) and 5.0 for the $512\times512$ resolution (340 epochs). The CFG scale is set to 3.0 for the model of 800 epochs at a $256\times256$ resolution. The guidance interval~\cite{interval_guidance} is set to 0.1 following previous work~\cite{wang2025pixnerd}.

\subsection{Text-to-Image Generation}
\label{appendix: t2i_train_details}
For text-to-image generation, we trained our model on the BLIP3o~\cite{blip3o} dataset, which contains approximately 36M pretraining images and 60k high-quality instruction-tuning data. We adopt Qwen3-1.7B~\cite{yang2025qwen3} as the text encoder. The entire training takes about 6 days on 8$\times$ H800 GPUs.
We adopt Qwen3-1.7B~\cite{yang2025qwen3} as the text encoder. To improve the alignment of frozen text features \cite{fluid}, we jointly train several transformer layers on the frozen text features similar to Fluid~\cite{fluid}. The total batch size is 1536 for $256\times256$ resolution pretraining and 512 for $512\times512$ resolution pretraining. Following PixNerd~\cite{wang2025pixnerd}, we pretrain DeCo on $256\times256$ resolution for 200K steps and pretrain on $512\times512$ resolution for 80K steps. We further fine-tune the pretrained DeCo on BLIP3o-60k with 40k steps at the $512\times512$ resolution following PixNerd. We adopt the gradient clip to stabilize training. \textit{The whole training only takes about 6 days on 8$\times$ H800 GPUs.} We use the Adams-2nd solver with 25 steps as the default choice for sampling. The cfg scale is set to 4.0. We leave the native resolution~\cite{nit} or native aspect training~\cite{seedream2, seedream3, mogao} as future works.

\subsection{Experiment Configurations}
Table \ref{appendix:tab_config} summarizes the experiment configurations for DeCo-L/16, DeCo-XL/16, and DeCo-XXL/16.
In practice, we follow the training setups from previous works such as DiT~\cite{dit}, SiT~\cite{sit}, and PixNerd~\cite{wang2025pixnerd}. Besides, we sweep the CFG scale within the given ranges using an interval of 0.1. 

\begin{table}[t]
\centering
\resizebox{1.0\width}{!}{
\tablestyle{6pt}{1.0}
\begin{tabular}{l |ccc}
 & \textbf{DeCo-L} & \textbf{DeCo-XL} & \textbf{DeCo-XXL} \\
\shline
\rowcolor[gray]{0.9}\multicolumn{4}{l}{\textbf{architecture}} \\
DiT depth & 22 & 28 & 16  \\
hidden dim & 1024 & 1152 & 1536  \\
heads & 16 & 16 & 24  \\
params & 426M & 682M & 1.1B  \\
decoder depth & &3& \\
decoder hidden dim & & 32& \\
patch size & &16 &\\
image size & \multicolumn{3}{c}{256 (other settings: 512)} \\
\midline
\rowcolor[gray]{0.9}\multicolumn{4}{l}{\textbf{training}} \\
optimizer & \multicolumn{3}{c}{AdamW~\cite{loshchilov2018decoupled}, $\beta_1, \beta_2=0.9, 0.999$} \\
batch size & \multicolumn{3}{c}{256} \\
learning rate & \multicolumn{3}{c}{1e-4} \\ 
lr schedule & \multicolumn{3}{c}{constant} \\
weight decay & \multicolumn{3}{c}{0} \\ 
ema decay & \multicolumn{3}{c}{0.9999} \\
time sampler & \multicolumn{3}{c}{$\text{logit}(t){\sim}\mathcal{N}(\mu, \sigma^2)$, $\mu$ = 0, $\sigma$ = 1 } \\
noise scale & \multicolumn{3}{c}{1.0} \\
\midline
\rowcolor[gray]{0.9}\multicolumn{4}{l}{\textbf{sampling}} \\
ODE solver & \multicolumn{3}{c}{Euler} \\
ODE steps & \multicolumn{3}{c}{100} \\
time steps & \multicolumn{3}{c}{linear in [0.0, 1.0]} \\
CFG scale range & \multicolumn{3}{c}{[3.0-3.2]~(256$\times$256), [4.5-5.0]~(512$\times$512)}  \\
CFG interval \cite{interval_guidance} & \multicolumn{3}{c}{{[0.1, 1]}} \\
\end{tabular}
}
\caption{{Configurations of experiments.}}
\label{appendix:tab_config}
\end{table}

\section{Text-to-Image Prompts}
\label{appendix:textual_prompts}

Below, we list the prompts used for text-to-image generation in \cref{fig:t2i_visualization}.
These prompts cover a mix of animals, people, and scenes to evaluate semantic understanding and visual detail generation.

\begin{minipage}{0.95\columnwidth}\vspace{0mm}    \centering
\begin{tcolorbox} 
    \centering
\begin{itemize}
\setlength{\itemsep}{2pt}
    \item A lovely horse stands in the bedroom.
    \item A baby cat stands on two legs, wearing a chothes.
    \item A cyberpunk woman with glowing tattoos and a mechanical arm beneath a holographic sky.
    \item  A man sipping coffee on a sunny balcony filled with potted plants, wearing linen clothes and sunglasses, basking in the morning light.
    \item A beautiful woman.
    \item A cute panda is wielding a sword in realistic style.
    \item An extremely happy American Cocker Spaniel is smiling and looking up at the camera with his head tilted to one side.
    \item A raccoon wearing a detective’s hat, observing something with a magnifying glass.
    \item Close-up of an aged man with weathered features and sharp blue eyes peering wisely from beneath a tweed flat cap.
\end{itemize}
\end{tcolorbox}
\end{minipage}

\section{Quantization Tables}
\label{appendix: sec_quantization_tab}

In our DeCo, we use the normalized reciprocal of scaled JPEG quantization tables as adaptive weights to emphasize different frequency components. These tables are a core component of the JPEG compression standard and are designed based on properties of the human visual system (HVS)~\cite{pennebaker1992jpeg}.

As shown in \cref{sec: method_freqfm_loss}, a quantization table is an 8$\times$8 matrix that determines the compression level for each frequency coefficient after the Discrete Cosine Transform (DCT). The JPEG standard uses two separate tables: one for the luminance (Y) component and another for the chrominance (Cb/Cr) components. 
This design is based on key characteristics of human perception.
The core principle is that the human eye is not equally sensitive to all visual information. Specifically, two HVS properties are crucial. Firstly, the human eye is much more sensitive to low-frequency components than to high-frequency components. Secondly, the eye is more sensitive to changes in brightness (luminance) than in color (chrominance)~\cite{pennebaker1992jpeg}.

\begin{figure}[t]
    \centering
    \includegraphics[width=1\linewidth]{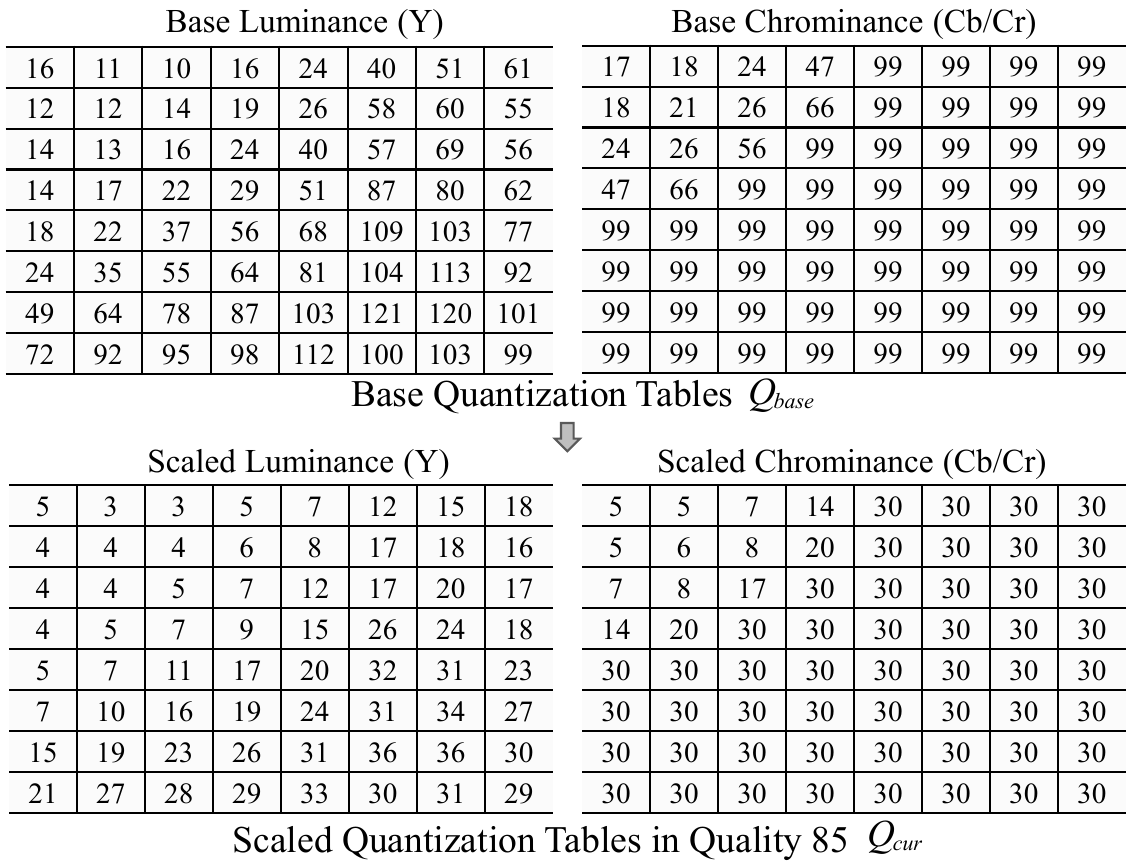}
    \caption{Base and Scaled Quantization Tables.}
    \label{fig:quantization_tables}
\end{figure}

Based on extensive experiments, the standard base quantization tables $Q_\text{base}$ in \cref{fig:quantization_tables} were developed to reflect these properties~\cite{pennebaker1992jpeg}. These tables have smaller values (finer quantization intervals) for low-frequency coefficients, which are perceptually more important. Conversely, these tables have larger values (coarser quantization intervals) for high-frequency coefficients, as the resulting information loss is less noticeable to the human eye. Similarly, the luminance table generally has smaller values than the chrominance table.
These base tables can be scaled using a quality factor $q$ to create new scaled quantization tables $Q_\text{cur}$ for different compression levels.

Since a smaller quantization step implies that a frequency component is more significant to human perception, we use the normalized reciprocal of the scaled quantization tables as adaptive weights, i.e., $\frac{1}{Q_\text{cur}}$ with normalization. This allows us to assign a higher weight to the frequency components that are visually more important in our frequency-aware flow-matching loss $\mathcal{L}_\mathrm{FreqFM}$.

\section{Pseudocodes for DeCo}
\definecolor{codeblue}{rgb}{0.25,0.5,0.5}
\definecolor{codekw}{rgb}{0.85, 0.18, 0.50}

\definecolor{codesign}{RGB}{0, 0, 255}
\definecolor{codefunc}{rgb}{0.85, 0.18, 0.50}
\definecolor{commentcolor}{rgb}{0.25,0.5,0.5}

\lstdefinelanguage{PythonFuncColor}{
  language=Python,
  keywordstyle=\color{black}\bfseries,
  commentstyle=\color{codeblue},
  stringstyle=\color{orange},
  showstringspaces=false,
  basicstyle=\ttfamily\small,
  literate=
    {dataloader}{{\color{codefunc}dataloader}}{1}
    {sample_t}{{\color{codefunc}sample\_t}}{1}
    {randn}{{\color{codefunc}randn}}{1}
    {randn_like}{{\color{codefunc}randn\_like}}{1}
    {jvp}{{\color{codefunc}jvp}}{1}
    {stopgrad}{{\color{codefunc}stopgrad}}{1}
    {l2_loss}{{\color{codefunc}l2\_loss}}{1}
    {kmeans}{{\color{codefunc}kmeans}}{1}
    {colormap}{{\color{codefunc}colormap}}{1}
    {plot}{{\color{codefunc}plot}}{1}
    {ZigZagIndices}{{\color{codefunc}ZigZagIndices}}{1}
    {DCT2D}{{\color{codefunc}DCT2D}}{1}
    {Sum}{{\color{codefunc}Sum}}{1}
    {max}{{\color{codefunc}max}}{1}
    {mean}{{\color{codefunc}mean}}{1}
    {len}{{\color{codefunc}len}}{1}
    {RGB2YCbCr}{{\color{codefunc}RGB2YCbCr}}{1}
    {patchify}{{\color{codefunc}patchify}}{1}
}

\lstset{
  language=PythonFuncColor,
  backgroundcolor=\color{white},
  basicstyle=\fontsize{8pt}{9.5pt}\ttfamily\selectfont,
  columns=fullflexible,
  breaklines=true,
  captionpos=b,
  mathescape=true 
}

\begin{algorithm}[t]
\caption{Training step}
\label{alg:code_train}
\begin{lstlisting}
# $\color{commentcolor} \theta_\text{DiT}$: DiT network
# $\color{commentcolor} \theta_\text{Dec}$: Pixel Decoder network
# $\color{commentcolor} x_0$: training batch
# y: class label or textual prompt
# $\color{commentcolor} Q_\text{cur}$: scaled quantization tables in quality 85.

# Prepare inputs
t = sample_t()
$x_1$ = randn_like($x_0$)
$x_t$ = (1-t)$x_0$ + t$x_1$ # original scale
$\bar{x}_t$ = patchify($x_t$, patch_size=16) # small-scale

# Prepare ground-truth velocities
$v_t$ = $x_1$ - $x_0$
$\mathbb{V}_t$ = DCT2D(RGB2YCbCr($v_t$))

# Generate low-frequency semantic condition 
c = $\theta_\text{DiT}$($\bar{x}_t$, t, y)
# Predict velocity conditioned on c
$v_{\theta}$ = $\theta_\text{Dec}$($x_t$, t, c)
$\mathbb{V}_{\theta}$ = DCT2D(RGB2YCbCr($v_{\theta}$))

# Compute Loss
FM_loss = mean($\left\| {v}_\theta - {v}_t \right\|^2$) 
w = 1/$Q_\text{cur}$
w = w / w.mean() # normalized adaptive weights
FreqFM_loss = mean(w * $\left\| \mathbb{V}_{\theta} - \mathbb{V}_t \right\|^2$) 
loss = FM_loss + FreqFM_loss + REPA_loss
\end{lstlisting}
\end{algorithm}

\begin{algorithm}[t]
\caption{K-Means Visualization in~\cref{fig:intro}~(c)}
\label{alg:kmeans_vis}
\begin{lstlisting}
# Feats: DiT outputs (T, H, W, C)
# I: generated images (T, H, W, 3)
# T: Sampling steps

for t in T:
    # Flatten spatial dimensions
    f = Feats[t].reshape(-1, C)
    
    # Cluster pixel-wise features
    labels = kmeans(f, n_clusters=8)
    
    # Map clusters to visualization
    vis = colormap(labels).reshape(H, W)
    
    plot(vis, I[t])       
\end{lstlisting}
\end{algorithm}

\begin{algorithm}[t]
\caption{DCT Spectral Analysis}
\label{alg:dct_analysis}
\begin{lstlisting}
# V: Predicted velocity (B * T, ori_H, ori_W, C)
# Feats: DiT outputs (B * T, H, W, C)

# Pre-compute frequency scan order
# The block size of DCT is set to 8
idx = ZigZagIndices(block_size=8) 

F_energy, V_energy = 0, 0
F_num_blocks, V_num_blocks =0, 0
for f, v in (Feats, V):
    # Split input into patches
    f_patches = Unfold(f, kernel_size=8)
    v_patches = Unfold(v, kernel_size=8)
    
    # 2D Discrete Cosine Transform
    f_freq = DCT2D(f_patches)
    f_energy = f_freq ** 2
    v_freq = DCT2D(v_patches)
    v_energy = v_freq ** 2
    
    # Accumulate energy following ZigZag order
    # Maps 2D (u,v) to 1D frequency index
    F_energy += Sum(f_energy.reorder(idx))
    F_num_blocks += len(f_patches)
    V_energy += Sum(v_energy.reorder(idx))
    V_num_blocks += len(v_patches)

# Log-scale Normalization
Feats_S = log(1 + F_energy / F_num_blocks)
Feats_S = Feats_S / max(Feats_S)
V_S = log(1 + V_energy / V_num_blocks)
V_S = V_S / max(V_S)

plot(Feats_S)
plot(V_S)
\end{lstlisting}
\end{algorithm}

\subsection{Training Step of DeCo}
In \cref{alg:code_train}, we provide the pseudocodes for the training step of DeCo. DeCo utilizes the DiT to specialize in low-frequency semantic modeling with downsampled small-scale inputs $\bar{x}_t$. Semantic cues $c$ are hence incorporated with a lightweight pixel decoder to reconstruct high-frequency signals. In other words, the pixel decoder takes the low-frequency semantics $c$ from DiT as condition and predicts pixel velocities $v_{\theta}$ with a high-resolution input $x_t$. This new paradigm hence frees the DiT to specialize in modeling semantics, and allows for more specialized details generation. 
To emphasize visually salient frequencies and suppress perceptually insignificant high-frequency components, we further introduce a frequency-aware Flow-Matching loss $\mathcal{L}_\mathrm{FreqFM}$ inspired by the JPEG~\cite{JPEG}. 
A REPA~\cite{repa} loss is used in both our Baseline and DeCo.

\subsection{K-Means Visualization in \cref{fig:intro}~(c)}
In \cref{alg:kmeans_vis}, we provide the pseudocodes for the K-Means visualization in \cref{fig:intro}~(c). The number of clusters in K-Means is set to 8. We uniformly select 4 timesteps from the sampling process and visualize the clustering results at these timesteps.

\subsection{DCT energy distribution in \cref{fig:frequency_plot}}
In \cref{alg:dct_analysis}, we provide the pseudocodes for DCT spectral analysis of \cref{fig:frequency_plot}. We apply an 8$\times$8 DCT to transfrom the DiT outputs and pixel velocities into frequency domain. Each 8$\times$8 patch yields 64 frequency coefficients, which are then converted into energy via a square operation. 
We reorder these energies from low to high frequency using standard zigzag indexing, where lower indices correspond to lower-frequency components. Finally, we apply log-scale normalization to rescale all energies to the range [0, 1] for comparison.
As demonstrated in \cref{fig:frequency_plot}, compared with baseline, DeCo suppresses high-frequency signals in DiT outputs while preserving strong high-frequency energy in pixel velocity, confirming effective frequency decoupling. The distribution is computed on 10K images across all diffusion steps, i.e., $B$=10,000 and $T$=100.

\section{More Visualizations}
\begin{figure*}[t]
    \centering
    \includegraphics[width=0.95\linewidth]{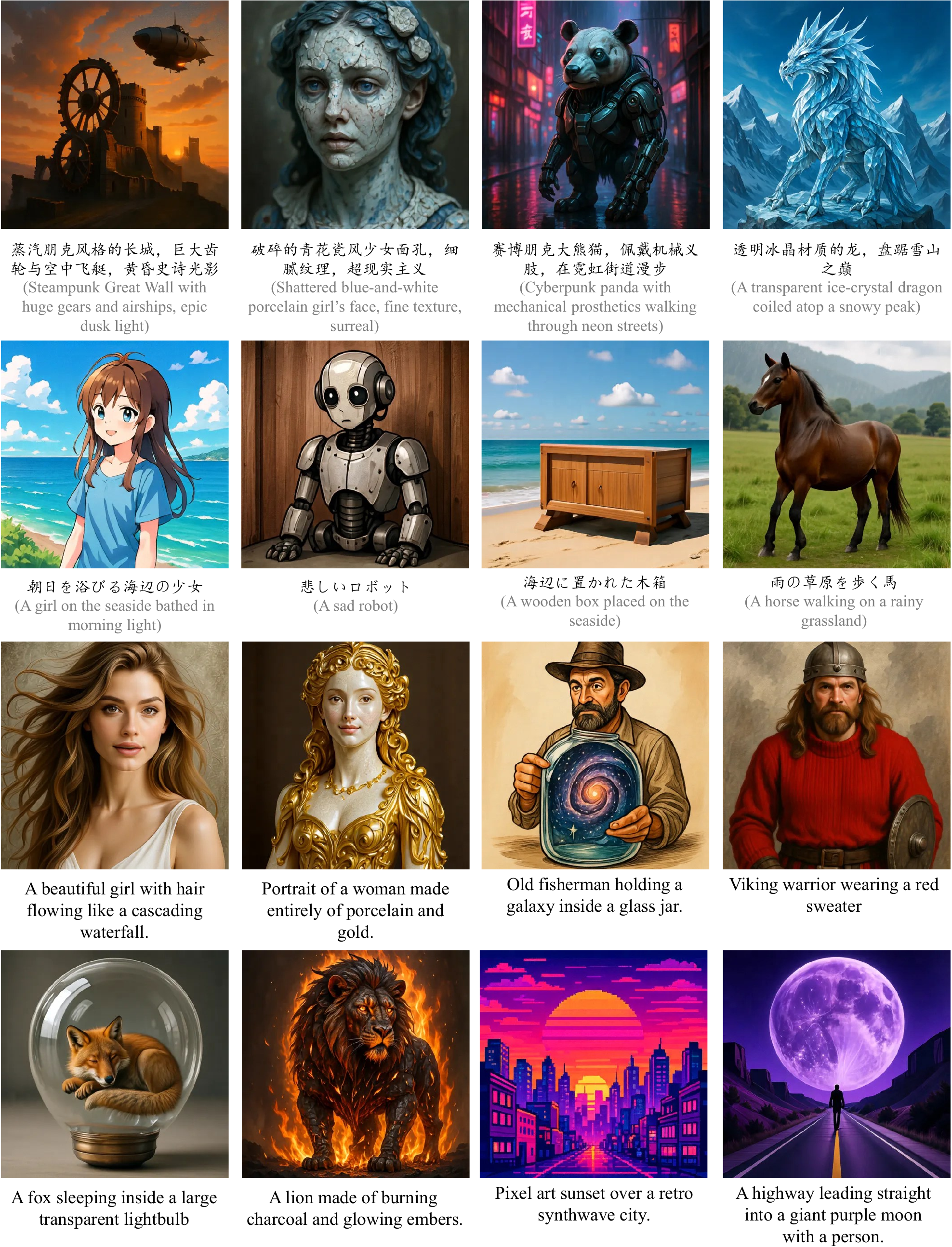}
    \caption{More Qualitative results of text-to-image generation at a 512$\times$512 resolution. Our DeCo supports multiple languages with the Qwen3 text encoder, such as Chinese, Japanese, and English.}
    \label{fig:visualization_512_t2i}
\end{figure*}

\begin{figure*}[t]
    \centering
    \includegraphics[width=0.95\linewidth]{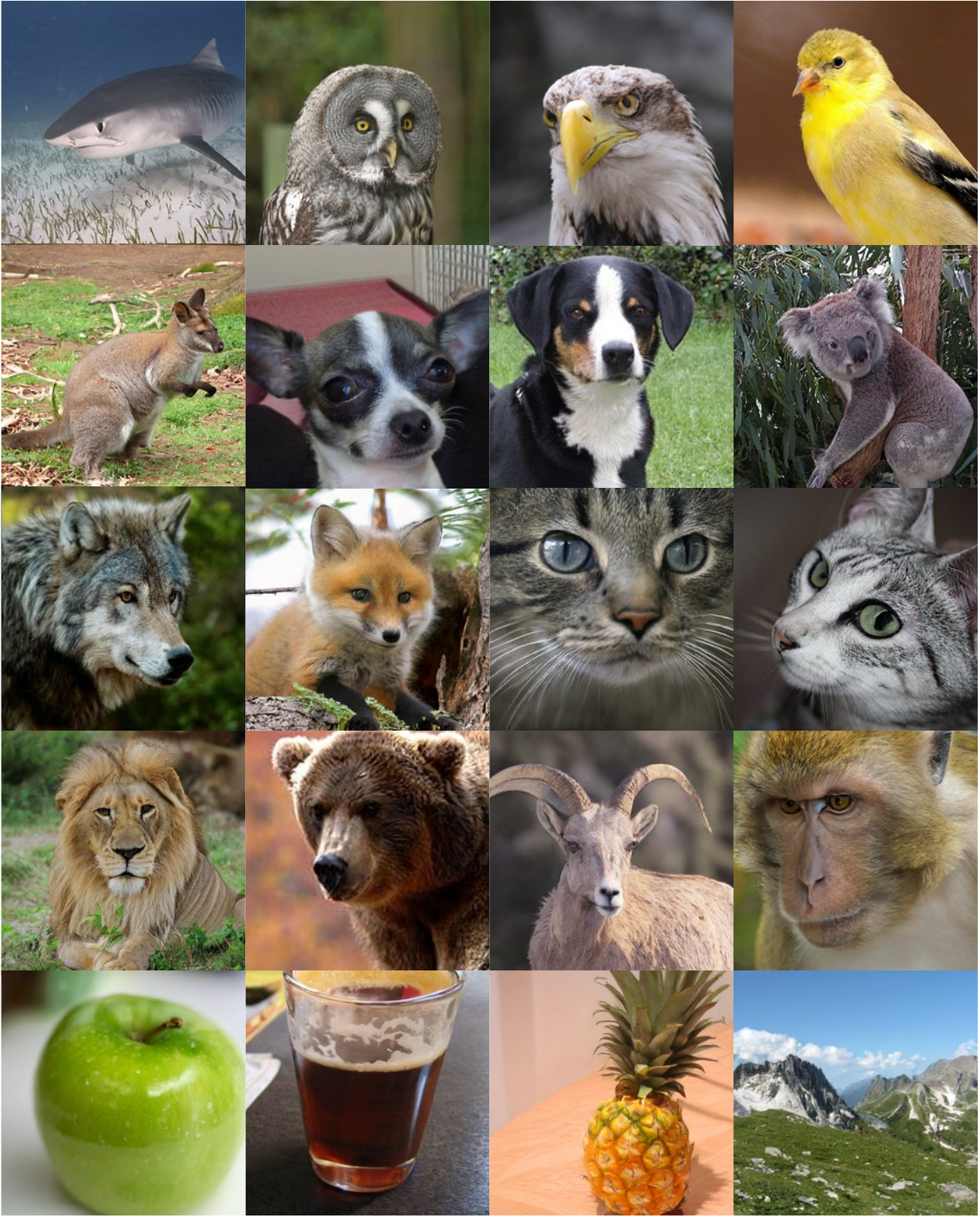}
    \caption{More qualitative results of class-to-image generation at a 256$\times$256 resolution.}
    \label{fig:visualization_256_imagenet}
\end{figure*}

\begin{figure*}[t]
    \centering
    \includegraphics[width=0.95\linewidth]{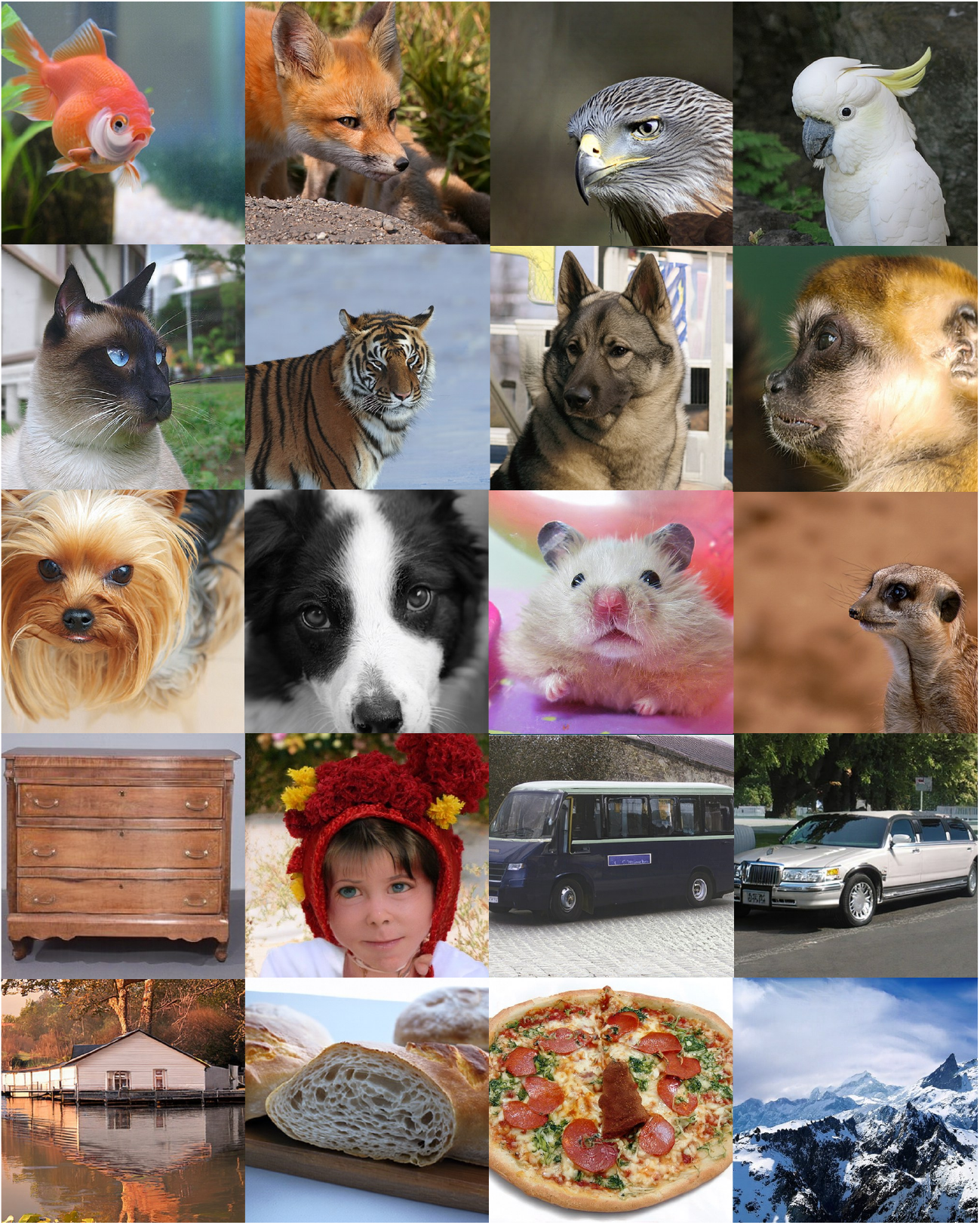}
    \caption{Qualitative results of class-to-image generation at a 512$\times$512 resolution.}
    \label{fig:visualization_512_imagenet}
\end{figure*}

In this section, we provide more visualizations, including text-to-image generation in \cref{fig:visualization_512_t2i}, class-to-image generation at a 256$\times$256 resolution in \cref{fig:visualization_256_imagenet}, and class-to-image generation at a $512\times 512$ resolution in \cref{fig:visualization_512_imagenet}. Our DeCo supports multiple languages with the Qwen3 text encoder after pretraining on the BLIP3o dataset~\cite{blip3o}, such as Chinese, Japanese, and English.

\end{document}